\pdfoutput=1

\documentclass[11pt]{article}

\usepackage{ACL2023}

\usepackage{times}
\usepackage{latexsym}

\usepackage[T1]{fontenc}

\usepackage[utf8]{inputenc}

\usepackage{microtype}

\usepackage{graphicx}
\usepackage{booktabs}
\usepackage{multicol}
\usepackage{multirow}
\usepackage{amsmath}
\usepackage{amssymb}
\usepackage{bbding}
\usepackage{subfigure}
\usepackage{colortbl}
\usepackage{pifont}
\usepackage{bbm}
\usepackage{makecell}
\usepackage{bbding}
\usepackage{mathrsfs}
\usepackage{bm}

\newcommand{\ourmethod}{\textsc{GanLM}}
\newcommand{\mourmethod}{\textsc{GanLM}-m}
\newcommand{\dtask}{replaced token detection}
\newcommand{\gtask}{replaced token denoising}
\newcommand{\Gtask}{Replaced Token Denoising}
\newcommand{\Dtask}{Replaced Token Detection}
\newcommand{\TLABEL}{\texttt{REPLACED}}
\newcommand{\FLABEL}{\texttt{ORIGINAL}}

\title{\ourmethod{}: Encoder-Decoder Pre-training with an Auxiliary Discriminator}

\author{
  Jian Yang\textsuperscript{\rm 1 \thanks{\ Contribution during internship at Microsoft Research Asia.}}, 
  Shuming Ma\textsuperscript{\rm 2},
  Li Dong\textsuperscript{\rm 2}, 
  Shaohan Huang\textsuperscript{\rm 2}, 
  {\bf Haoyang Huang}\textsuperscript{\rm 2}, 
  \\{\bf Yuwei Yin}\textsuperscript{\rm 3}, 
  {\bf Dongdong Zhang}\textsuperscript{\rm 2}, 
  {\bf Liqun Yang}\textsuperscript{\rm 1 \thanks{\ Corresponding author.}}, 
  {\bf Furu Wei}\textsuperscript{\rm 2}, 
  {\bf Zhoujun Li}\textsuperscript{\rm 1}\\
  \textsuperscript{\rm 1}State Key Lab of Software Development Environment, Beihang University \\ 
  \textsuperscript{\rm 2}Microsoft Research Asia; \textsuperscript{\rm 3}The University of Hong Kong \\
  \{jiaya, lqyang, lizj\}@buaa.edu.cn; \\ 
  \{shumma, lidong1, shaohanh, haohua, dozhang, fuwei\}@microsoft.com; \\
  yuweiyin@hku.hk
}

\begin{document}
\maketitle
\begin{abstract}
Pre-trained models have achieved remarkable success in natural language processing (NLP). However, existing pre-training methods underutilize the benefits of language understanding for generation. Inspired by the idea of Generative Adversarial Networks (GANs), we propose a GAN-style model for encoder-decoder pre-training by introducing an auxiliary discriminator, unifying the ability of language understanding and generation in a single model. Our model, named as \ourmethod{}, is trained with two pre-training objectives: \dtask{} and \gtask{}. Specifically, given masked source sentences, the generator outputs the target distribution and the discriminator predicts whether the target sampled tokens from distribution are incorrect. The target sentence is replaced with misclassified tokens to construct noisy previous context, which is used to generate the gold sentence. In general, both tasks improve the ability of language understanding and generation by selectively using the denoising data. Extensive experiments in language generation benchmarks show that \ourmethod{} with the powerful language understanding capability outperforms various strong pre-trained language models (PLMs) and achieves state-of-the-art performance.\footnote{\url{https://github.com/CSJianYang/GanLM}}

\end{abstract}

\section{Introduction}
The pre-training-then-fine-tuning paradigm has been proven successful in many natural language processing tasks~\cite{bert,roberta,gpt3}. While there are various pre-training approaches for the encoder-only architectures~\cite{electra,xlmr}, the encoder-decoder pre-training is underexplored, which is essential for natural language generation. To pre-train the entire encoder-decoder model, BART~\cite{bart} proposes a denoising language model objective and T5~\cite{t5} pre-trains the models with a span corruption objective. Furthermore, mBART \cite{mbart} and mT5 \cite{mt5} extend them to be multilingual pre-trained language models.

Unlike most encoder-decoder pre-training methods that simply apply sequence-to-sequence tasks on a single encoder-decoder architecture, we explore the approaches to pre-train the model in a GAN-style manner with an auxiliary discriminator. GAN~\cite{gan} performs well on both text and image generation tasks by combining the generator and discriminator. It aims to improve the ability of the generator to produce high-quality samples, which is important for the encoder-decoder pre-training when transferred to downstream generation tasks.
Similarly, MaskGAN~\cite{maskgan} shows the GAN-like training can improve the quality of the autoregressive language model. Therefore, it is intuitive to leverage GAN to empower the encoder-decoder pre-training by unifying language understanding and generation.

\begin{figure}[t]
\centering
\includegraphics[width=1.0\linewidth]{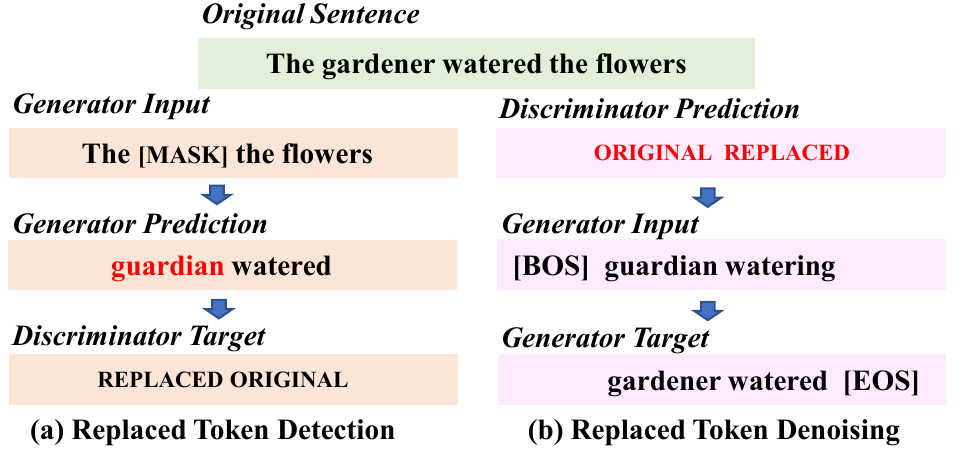}
\caption{A pre-training sample of our method, where \dtask{} (discriminator) and \gtask{} (generator) are used for pre-training. The discriminator classifies each generated token into \TLABEL{} or \FLABEL{}, where \TLABEL{} denote the predicted token is different from the gold token. The red fonts denote incorrect predictions. }
\label{intro}
\end{figure}

In this work, we propose a pre-training framework \ourmethod{}, using GAN-style learning to improve the transferability of pre-trained language models for the natural language generation.
Specifically, the encoder reads the masked source sentence and the generator obtains target distribution. 
Then, the discriminator distinguishes whether each token sampled from the target distribution matches the target gold sentence (\dtask{}). The misclassified tokens by discriminator are regarded as hard tokens for the generator to predict accurately. We replace original tokens in the target sentence with misclassified sampled ones to construct the noisy previous context for predicting the target sentence (\gtask{}).
In Figure \ref{intro}, the generator predicts the masked words ``guardian watered'', where the incorrect token ``guardian'' and correct token ``watered'' are both misclassified into \TLABEL{} and \FLABEL{} by the discriminator. Next, we resample a different token ``watering'' from the generated distribution.
Consequently, the target tokens ``gardener watered'' are replaced with the sampled tokens ``guardian watering'' to construct the noisy sample. The generator predicts the next word conditioned on previous noisy tokens (\gtask{}). Through combing two tasks, \ourmethod{} strengthen generation performance with the enhanced language understanding capability from the \dtask{} task.

Our method is effective for text generation and can be extended to natural language understanding tasks. We pre-train \ourmethod{} model on large-scale monolingual corpora and evaluate the performance of our pre-trained English model \ourmethod{} and multilingual model \mourmethod{} on various downstream tasks, including text summarization, machine translation, and data-to-text generation.
Experimental results demonstrate that our method substantially outperforms previous pre-trained encoder and sequence-to-sequence models on generation tasks. Our method is further tested on GLUE \cite{glue} and XNLI \cite{xnli} to validate the transferability of our pre-trained model. Analytic experiments emphasize the importance of the discriminator in both the pre-training and fine-tuning stage, leading to better performance.

\begin{figure*}[t]
\begin{center}
	\includegraphics[width=0.95\textwidth]{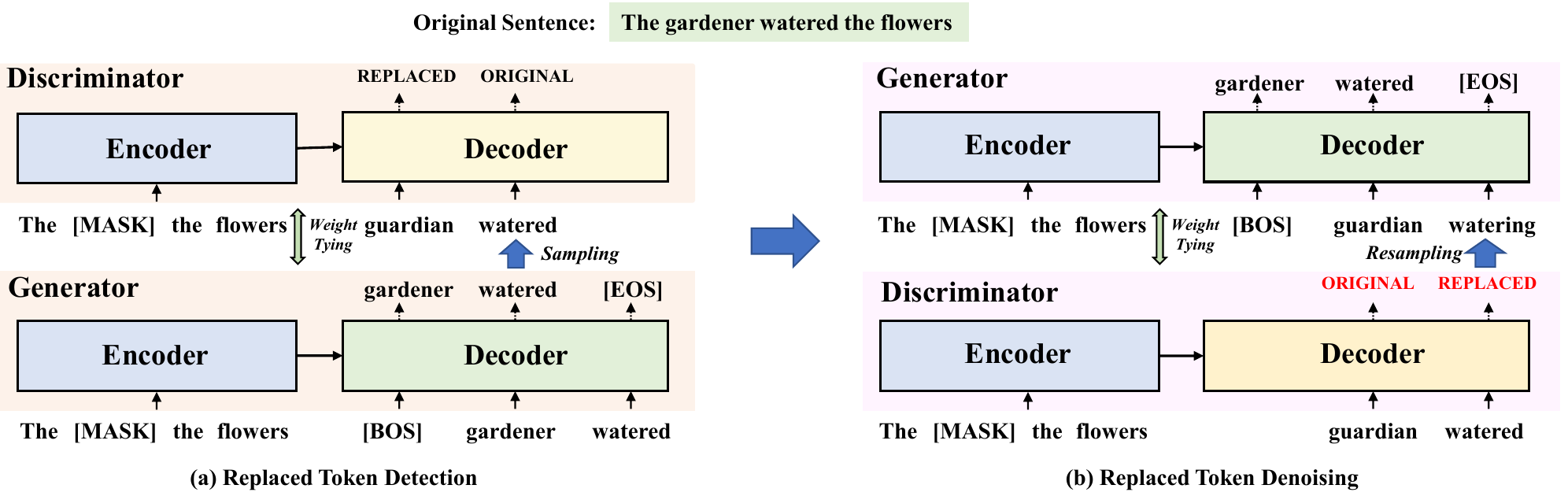}
	\caption{Overview of \ourmethod{}, including (a) \dtask{} and (b) \gtask{}. The encoder reads the source sentence and the generator obtains target distribution, where the generator and discriminator are supervised by the gold labels in (a). The discriminator distinguishes whether the sampled tokens ``guardian watered'' are replaced (both tokens are misclassified in this example). For the correct predicted token ``watered'', we obtain a different token ``watering'' by resampling. The target tokens are replaced with the misclassified tokens to construct the noisy input, which are used to predict the gold sentence ``gardener watered \texttt{[EOS]}'' in (b).}
	\label{model}
\end{center}
\end{figure*}

\section{\ourmethod{}}
\subsection{Model Overview}
Our GAN-style pre-trained model comprises a generator ($\mathcal{G}$) and discriminator ($\mathcal{D}$), which are both encoder-decoder frameworks and conditioned on the same encoder (Enc).
In Figure~\ref{model}, the encoder reads the masked sentence and the generator decoder obtains the target distribution. Then the discriminator decoder distinguishes whether each token in the sampled target sentence matches the gold reference. Tokens in the target gold sentence are randomly replaced with misclassified ones by the discriminator to construct the noisy sample, which is fed into the generator decoder to predict the target sentence (\gtask{}).

\subsection{Masked Sequence Generator}
Given a monolingual sentence $x=(x_1,\dots,x_n)$ with $n$ words from the dataset $D_{k}$ of language $L_k \in L_{all}=\{L_1,\dots,L_K\}$ $(\lvert L_{all} \rvert=K)$, some random spans of contiguous tokens in $x$ are corrupted as the source sentence, which is denoted as $x^{src}=(x_1,\dots,x_{\setminus u:v},\dots,x_n)$. $x_{\setminus u:v}$ is a masked span of $x_{u:v}$, where the fragment from position $u$ to $v$ is corrupted by \texttt{[MASK]}. Given $x^{src}$, the generator predicts the original identities of the masked tokens $x^{trg}=(x_{\setminus 1},\dots,x_{u:v},\dots,x_{\setminus n})$ autoregressively:
\begin{MiddleEquation}
\begin{equation}
    x^{trg}_t = \text{Enc-Dec}(x^{src},x^{trg}_{1:t-1};\{\theta_{\mathcal{E}},\theta_{\mathcal{G}}\})
    \label{generator}
\end{equation}
\end{MiddleEquation}where $\theta_{\mathcal{E}}$ and $\theta_{\mathcal{G}}$ denote the encoder and decoder parameters of the generator. Enc-Dec denotes an encoder-decoder model. The generator predicts the next position $t$ token $x^{trg}_{t}$ based on previous tokens.

The training objective of sequence-to-sequence masked language modeling (S2S-MLM) on the dataset $D_{k}$ of language $L_k$ is defined as:
\begin{MiddleEquation}
\begin{equation}
    \mathcal{L}_{\mathcal{G}} = \mathbb{E}_{x \sim D_{k}} \left[ \log P_{G}(x^{trg}|x^{src}; \{\theta_{\mathcal{E}},\theta_{\mathcal{G}}\}) \right]
    \label{generator_objective}
\end{equation}
\end{MiddleEquation}where $x^{src}$ and $x^{trg}$ are derived from $x$.

\subsection{Replaced Token Detection}
The generator outputs the distribution of each target token and we create a sampled sentence $\hat{x}^{trg}$ by randomly sampling tokens from the distribution. The discriminator distinguishes whether each token in $\hat{x}^{trg}$ is replaced compared to $x^{trg}$. 
Given the target distribution $P_{G}(x^{trg}_{t}|x^{src})$ $(x^{trg}_{t} \in x^{trg})$ from the generator, we construct $\hat{x}^{trg}$ for the discriminator:
\begin{MiddleEquation}
\begin{align}
\begin{split}
    \hat{x}^{trg} &= \textsc{replace}(x^{trg};x'_t) 
    \\ w.r.t. \; x'_{t} &\sim P_{G}(x^{trg}_{t}|x^{src}) \land x^{trg}_{t} \in x^{trg}
    \label{discriminator_input}
\end{split}
\end{align}
\end{MiddleEquation}where $\textsc{replace}(\cdot)$ replaces target $t$-th position unmasked token in $x^{trg}$ with the sampled token $x_{t}'$ from the generated distribution $P_{G}(x^{trg}_{t}|x^{src})$.

Given the source sentence $x^{src}$ and the encoder $\theta_{\mathcal{E}}$, the decoder of the discriminator $\theta_{\mathcal{D}}$ obtains a sequence of hidden representations $H_d=(h_1,\dots,h_n)$ by feeding the sampled sentence $\hat{x}^{trg}$ to the discriminator decoder:
\begin{MiddleEquation}
\begin{equation}
    H_d = \text{Enc-Dec}(x^{src},\hat{x}^{trg};\{\theta_{\mathcal{E}},\theta_{\mathcal{D}}\})
    \label{objective}
\end{equation}
\end{MiddleEquation}where $\theta_{\mathcal{E}}$ and $\theta_{\mathcal{D}}$ denote the encoder and decoder parameters of the discriminator. The decoder of the discriminator $\theta_{\mathcal{D}}$ adopts the bidirectional language model to classify each input token by extracting the past and future representations.
 
Given the representations $H_d$, the discriminator classifies sampled tokens $\hat{x}^{trg}$ into the \TLABEL{} or \FLABEL{} label with a sigmoid function $\sigma$:
\begin{MiddleEquation}
\begin{align}
\begin{split}
V = \sigma(H_d W_d)
\end{split}
\end{align}
\end{MiddleEquation}where $W_d \in R^{d_{e} \times 2}$ is the matrix projects the token representations to two categories (\TLABEL{} or \FLABEL{}) and $d_{e}$ is the model hidden size.

The training objective of the \dtask{} task for the discriminator is:
\begin{TinyEquation}
\begin{align}
\begin{split}
\mathcal{L}_{\mathcal{D}} = \mathbb{E}_{x \sim D_{k}} [\mathbbm{1}(\hat{x}^{trg} = x^{trg})\log V + \mathbbm{1}(\hat{x}^{trg} \neq x^{trg})\log(1 - V)]
\end{split}
\end{align}
\end{TinyEquation}where $\mathbbm{1}(\cdot)$ is the indicator function.
\subsection{\Gtask{}}
Although our model structure is similar to GAN, the generator is trained with maximum likelihood rather than the standard GAN objective due to the difficulty of the GAN training in NLP. We replace tokens in $x^{trg}$ with misclassified tokens by discriminator to construct the noisy previous context $x^{trg}_{f}$. If the sampled token $\hat{x}^{trg}_{t}=x_{t}$ is labeled with \FLABEL{}, we will resample the token $x_{t}'$ ($x_{t}' \neq x_{t}$) from target distribution as the misclassified token to modify $x_{t}$ in $x^{trg}$. When $\hat{x}^{trg}_{t}=x_{t}'$ ($x_{t}' \neq x_{t}$) is labeled with \TLABEL{}, the miscalssified token $x_{t}'$ directly replaces $x_{t}$ in the target sentence. Given the target sentence $x^{trg}$ and generated probabilities $P_{G}$, we replace tokens in $x^{trg}$ with sampled tokens as the previous noisy context:
\begin{SmallEquation}
\begin{align}
\begin{split}
 x^{trg}_{f} &= \textsc{replace}(x^{trg};\hat{x}^{trg}_t)\\
 w.r.t. \; \hat{x}^{trg}_{t} &\sim P_{G}(x^{trg}_{t}|x^{src}) \land t \in v
\end{split}
\end{align}
\end{SmallEquation}where $v=\{v_1,\dots,v_p\}$ $(\lvert v \rvert = p)$ denotes the positions in $x^{trg}$ of the misclassified tokens. 

The training objective of the \gtask{} ($\mathcal{DG}$) task based on the source sentence $x^{src}$ and target noisy context $x^{trg}_{f}$ is described as:
\begin{SmallEquation}
\begin{equation}
    \mathcal{L}_{\mathcal{DG}}= \mathbb{E}_{x \sim D_{L_k}} [ -\log P(x^{trg}|x^{src},x^{trg}_{f}; \{\theta_{\mathcal{E}},\theta_{\mathcal{D}}\}) ]
    \label{noisy_objective}
\end{equation}
\end{SmallEquation}where $x^{trg}$ is predicted by the previous noisy tokens $x^{trg}_{f}$ instead of previous gold context.

\subsection{Multi-task Learning}
Given multilingual corpora $D_{all}=\{D_{1},\dots,D_{K}\}$ of $K$ languages, the pre-trained model with parameters $\{\theta_{\mathcal{E}},\theta_{\mathcal{G}},\theta_{\mathcal{D}}\}$ is jointly trained over $K$ languages to optimize the combined self-supervised objective as below: 
\begin{MiddleEquation}
\begin{align}
\begin{split}
    \mathcal{L}_{\mathcal{P}} = \mathbb{E}_{L_k \in L_{all}}[\mathcal{L}_{\mathcal{G}} + \lambda \mathcal{L}_{\mathcal{D}} + \mathcal{L}_\mathcal{{DG}}]
    \label{pretrain-objective}
\end{split}
\end{align}
\end{MiddleEquation}where $\lambda=10.0$ is the discriminator weight and $L_{all}=\{L_1,\dots,L_K\}$. To improve model efficiency, a tiny discriminator decoder (4 layers) is adopted to help generator decoder (12 layers).

\section{Discriminator-enhanced Fine-tuning}
To fully utilize the pre-trained parameters, we keep the auxiliary discriminator in downstream generation tasks (discriminator-enhanced fine-tuning) to enhance the generator, where both the pre-trained generator and discriminator are recycled. Given the annotated corpus $D_{s}$ of $K$ languages, the pre-trained model $\{\theta_{\mathcal{E}},\theta_{\mathcal{D}},\theta_{\mathcal{G}}\}$ is optimized by: 
\begin{MiddleEquation}
\begin{align}
\begin{split}
    \mathcal{L}_{\mathcal{F}} = \mathbb{E}_{x,y \sim D_{s}}[\mathcal{L}_{\mathcal{G}} + \lambda \mathcal{L}_{\mathcal{D}} + \mathcal{L}_{\mathcal{DG}}]
    \label{fine-tune-objective}
\end{split}
\end{align}
\end{MiddleEquation}where $x$ and $y$ are the parallel pair from $D_{s}$. The objective in the fine-tuning stage use the original pair $x$ and $y$ without S2S-MLM. The generator $\{\theta_{\mathcal{E}},\theta_{\mathcal{G}}\}$ are kept for inference by throwing out the discriminator decoder $\theta_{\mathcal{D}}$. Alternatively, the discriminator $(\mathcal{D}$: $\{\theta_{\mathcal{E}},\theta_{\mathcal{D}}\})$ or generator $(\mathcal{G}$:$\{\theta_{\mathcal{E}},\theta_{\mathcal{G}}\})$ can also be separately fine-tuned on the downstream task. 

\begin{table*}[t]
\begin{center}
\small
\resizebox{1.0\textwidth}{!}{
\begin{tabular}{lllcccc}
\toprule
\toprule
\textbf{ID} &\bf Model & \bf Pre-training Objective &\bf Summarization   &   \multicolumn{3}{c}{\bf Translation}  \\
&& & RG-1/RG-2/RG-L & Avg$_{En \to X}$ & Avg$_{X \to En}$ & Avg$_{all}$   \\
\midrule
{\large{\ding{172}}} & Transformer w/o Pretraining  &- & 32.36/11.46/25.47 & 21.4  & 25.5  & 23.5 \\ 
\midrule
{\large{\ding{173}}} & BERT/mBERT~\cite{bert}              & Masked Language Model    & 36.93/15.00/29.62       & 26.4  & 29.6  & 28.0 \\ 
{\large{\ding{174}}} & ELECTRA~\cite{electra}              & Replaced Token Detection & 43.02/19.94/34.83       & 29.1  & 32.8  & 30.3 \\ 
{\large{\ding{175}}} & BART~\cite{bart}/mBART~\cite{mbart} & Denoising Autoencoder    & 44.13/21.04/36.02       & 30.3  & 33.3  & 31.4 \\
{\large{\ding{176}}} & T5~\cite{t5}/mT5~\cite{mt5}         & Span Corruption          & 44.22/21.06/36.12       & 30.4  & 33.6  & 31.7 \\
\midrule
{\large{\ding{177}}} &\bf \ourmethod{}/\mourmethod{} (ours)       & \Dtask{} + \Gtask{}        & \bf 45.36/21.98/36.84 & \bf 31.2 & \bf 34.2  & \bf 32.8
\\
{\large{\ding{178}}} & {\large{\ding{177}}} - Discriminator-enhanced Fine-tuning  & \Dtask{} + \Gtask{}        & 44.74/21.47/36.40   & 31.1 & 34.0 & 32.6 \\ 
{\large{\ding{179}}} & {\large{\ding{178}}} - \Gtask{}                     &  \Dtask{}                  & 44.28/21.14/36.24   & 30.6 & 33.6 & 32.1 \\ 
\bottomrule
\bottomrule
\end{tabular}
}
\end{center}
\caption{\label{task_study} Comparison of different pre-training objectives. Particularly, all methods in this table use the base-setting model and are pre-trained with 500K steps on the same corpora for a fair comparison. We report ROUGE scores for abstractive text summarization (XSum) and BLEU scores for multilingual machine translation (IWSLT-17). }
\end{table*}

\begin{table}[t]
\centering
\resizebox{1.0\columnwidth}{!}{
\small
\begin{tabular}{lllcc}
\toprule
{\bf Model}  & {\bf \#Corpus} & {\bf XSum} & {\bf CNN / DailyMail} \\
 & & { RG-1/RG-2/RG-L} & { RG-1/RG-2/RG-L} \\ \midrule
\textsc{PtrNet}~\cite{ptrnet} &  - & 28.10/8.02/21.72  &  39.53/17.28/36.38\\ 
\midrule
{MASS}~\cite{mass}                             & -     &  39.75/17.24/31.95    & 42.12/19.50/39.01 \\
\textsc{BERTSumAbs}~\cite{bertsum}             & 16GB  &  38.76/16.33/31.15    & 41.72/19.39/38.76 \\
RoBERTa  \cite{roberta}                        & 160GB & 42.19/19.22/34.23     & 41.28/19.11/38.57 \\
{\textsc{ERNIE-GEN}}~\cite{erniegen}           & 16GB  &  -                    & 42.30/19.92/39.68  \\
T5 \cite{t5}                                   & 750GB &  -                    & 42.05/20.34/39.40 \\
UniLM      \cite{unilm}                        & 16GB  & -                     & 43.08/20.43/40.34 \\
UniLMv2  \cite{unilmv2}                        & 160GB & 44.00/21.11/36.08     & 43.16/20.42/40.14 \\
RoBERTa + \textit{s2s-ft} \cite{s2s_ft}        & 160GB & 43.39/20.55/35.63     & 42.28/20.21/39.87 \\
UniLMv2 + \textit{s2s-ft} \cite{s2s_ft}        & 160GB & 44.37/21.54/36.61     &  43.89/21.05/41.02\\
\bf \ourmethod{} (ours)                        & 160GB & \bf 45.36/21.98/36.84 & \bf 44.15/21.12/41.32  \\
\bottomrule
\end{tabular}
}
\caption{
Abstractive summarization results on the test set of CNN / DailyMail, and XSum. The evaluation metric is the F1 score of ROUGE (RG) scores. }
\label{tab:cnndm_xsum}
\end{table}

\section{Experiment Setting}
\subsection{Pre-training Details}
\paragraph{Model Configuration}
In the experiments, we adopt a sequence-to-sequence base-setting Transformer architecture with 768 hidden size, 3072 FFN (feed-forward network) dimension, 12 attention heads, and 12 encoder/decoder layers.  The maximum sequence length of learned positions embeddings in the encoder/decoder is set as 1024. All token embedding matrices and output projection matrix parameters are shared for model efficiency.

\paragraph{Dataset}
Following the previous work \cite{roberta}, our English pre-trained model \ourmethod{} is trained on 160GB English monolingual data from BookCorpus, CC-News, OpenWebText, and CC-Stories. In addition, we pre-train \mourmethod{} with 6TB multilingual data as the pioneering work \cite{deltalm}, which is a combination of CC100, CC-Net, and Wikipedia, covering 100 languages. All texts are tokenized by SentencePiece \cite{sentencepiece} and encoded by the dictionary from XLM-R \cite{xlmr}.

\paragraph{Optimization}
For S2S-MLM, we randomly mask 15\% of the words in each instance with an average span length of 3 \cite{t5}. For the \dtask{}, we set the discriminator weight $\lambda=10.0$. We adopt Adam \cite{adam} with a learning rate of 3e-4 and 10K warm-up steps for pre-training. The model is trained on 128 NVIDIA A100 GPUs (40GB) from scratch and each batch contains 8K samples. The English pre-trained model \ourmethod{} and multilingual model \mourmethod{} are trained for 500K steps. Specifically, all methods in Table \ref{task_study} are pre-trained with 500K steps for a fair comparison.

\subsection{Downstream Tasks}
\paragraph{Monolingual Summarization} \textbf{CNN / DailyMail} \cite{ptrnet} is an abstractive summarization dataset aiming at generating a concise summary from an English news article in CNN and DailyMail.
As a popular abstractive summarization dataset, \textbf{XSum} \cite{xsum} compresses a BBC news article to a short one-sentence summary.
\paragraph{Multilingual Summarization} To test the capability of our multilingual pre-trained model, a large-scale multilingual dataset named \textbf{WikiLingua} \cite{wikilingua} of 18 languages from WikiHow is used to evaluate multilingual abstractive summarization systems.

\paragraph{Bilingual Translation} For the bilingual task, we use the \textbf{WMT-14 English-German}, \textbf{WMT-14 English-French}, and \textbf{WMT-16 English-Romanian} dataset for evaluation. WMT-14 En-De from WMT consists of 4.5M sentence pairs and the newstest2014 is used as the test set. WMT-14 En-Fr is a large-scale dataset containing nearly 41M sentence pairs and newstest2014 is adopted for evaluation. WMT-16 En-Ro is comprised of original parallel sentences and back-translation data. 

\paragraph{Multilingual Translation} \textbf{IWSLT-17} of 5 languages and \textbf{WMT-10} of 11 languages are utilized for multilingual translation. For IWSLT-17, English (En), German (De), Italian (It), Dutch (Nl), and Romanian (Ro) corpora are downloaded from the IWSLT-2017 benchmark. We use dev2010 for validation and tst2017 for test. For WMT-10, we use the parallel data of 11 languages from the WMT benchmark for evaluation \cite{zcode}.

\paragraph{Data-to-Text Generation} Data-to-text generation accepts multiple triplets and produces a description. WebNLG \cite{WebNLG} contains parallel DBpedia triple sets and short texts. The En-En direction contains 17K triple sets and 45K short texts and the En-Ru direction contains 7K triple sets and 19K texts in Russian. The ROUGE scores on the valid set are reported for a fair comparison with the previous work \cite{gem}.

\subsection{Fine-tuning Details}
\paragraph{Abstractive Summarization} 
During fine-tuning, we use the Adam \cite{adam} optimizer with an initial learning rate of 1e-4 and the batch size is set as 2048 tokens on 8 V100 GPUs. The models are trained with the label smoothing cross-entropy with a smoothing ratio of 0.1.

\paragraph{Neural Machine Translation} For the large-scale multilingual dataset WMT-10, our pre-trained model is fine-tuned on 32 V100 GPUs with a learning rate of 3e-4. For all bilingual translation tasks and the IWSLT-2017 benchmark, we adopt Adam with a learning rate of 1e-4 and set the batch size as 2048 tokens on 8 V100 GPUs.

\paragraph{Data-to-text Generation} 
We use Adam with a learning rate of \{8e-5,1e-4\} and set the batch size as 16 sentences on the WebNLG dataset.

\begin{table}[t]
\begin{center}
    \resizebox{1.0\columnwidth}{!}{
    \begin{tabular}{lccc}
    \toprule
    \bf Model   & \bf En & \bf Zh  & {\bf Avg$_{\bm{18}}$}  \\
        \midrule
        Transformer \cite{transformer} & 35.9/13.3/29.6 & 32.1/16.2/26.6  & 29.9/10.7/25.0 \\
        XLM-R \cite{xlmr}              & 41.4/17.6/34.5 & 42.2/23.8/34.9  & 37.5/16.0/31.2  \\
        mBART \cite{mbart}             & 44.2/20.0/32.1 & 44.8/25.8/37.6  & 40.1/18.2/33.7 \\
        \bf \mourmethod{} (ours) & \bf 44.7/20.6/37.8 &  \bf 45.7/26.4/38.0 & \bf 40.5/18.6/34.0 \\
        \bottomrule
    \end{tabular}}
    \caption{Results of our method and other baselines on multilingual abstractive summarization. We report the RG-1/RG-2/RG-L (ROUGE) F1 scores of the 18 WikiLingua languages and the average scores.
    \label{tab:wikilingua}}
\end{center}
\end{table}

\section{Comparing Pre-training Objectives}
To verify the potential of our pre-training task under a fair comparison, we re-implement previous pre-training tasks and pre-trains baselines on the same corpora with 500K steps, including BERT/mBERT \cite{bert}, ELECTRA \cite{electra}, BART \cite{bart}/ mBART \cite{mbart}, and T5 \cite{t5}/mT5 \cite{mt5}. Table \ref{task_study} reports the ROUGE and BLEU points on the summarization dataset XSum and multilingual translation dataset IWSLT-17. All models have 12 encoder and 12 decoder layers with a hidden size of 768.
We observe that the encoder-decoder pre-trained model (T5/mT5) outperforms the pre-trained encoder (ELECTRA, BERT/mBERT), which corroborates the encoder-decoder pre-training is more beneficial to the downstream generation task. Experiments {\large{\ding{177}}}$\sim${\large{\ding{179}}} show the 
importance of the discriminator and \gtask{}. Experiment {\large{\ding{179}}} demonstrates that only the \dtask{} task can still bring improvement through strengthening the encoder shared by both generator and discriminator. Besides, the \dtask{} task is also helpful to downstream language understanding tasks with a powerful encoder. Lastly, the results verify that fine-tuning with the help of the pre-trained auxiliary discriminator further improves performance.

\section{Results of \ourmethod{}}
The English pre-trained model \ourmethod{} is evaluated on the abstractive text summarization task with the ROUGE \cite{rouge} scores.

\paragraph{XSum}
As shown in Table \ref{tab:cnndm_xsum}, the pre-training methods achieve significant improvements over the strong baseline \textsc{PtrNet} without pre-training. The sequence-to-sequence pre-trained model such as UniLMv2 + \textit{s2s-ft} outperforms other pre-training baselines, where the pseudo-masked technique is applied to the fine-tuning stage. Our method beats all pre-training baselines by a large margin with the discriminator-enhanced fine-tuning strategy. It emphasizes the importance of the fine-tuning strategy for the performance of downstream tasks. 

\paragraph{CNN / DailyMail} Our method is also evaluated on the CNN / DailyMail dataset in Table \ref{tab:cnndm_xsum}. The comparisons further indicate that our method obtains strong performance on generation by leveraging the discriminator.

\section{Results of \mourmethod{}}
To evaluate the multilingual pre-trained model \mourmethod{}, we report the BLEU \cite{bleu} scores for machine translation and ROUGE \cite{rouge} scores for text summarization and data-to-text generation.

\paragraph{WikiLingua} Table \ref{tab:wikilingua} reports the average ROUGE scores of 18 WikiLingua languages. The large improvement over other pre-training method demonstrate the summarization ability of our \mourmethod{}.

\paragraph{WMT14 En-De} The results on the bilingual translation are presented at Table \ref{tab:wmt14-en-de}. We observe that the proposed \ourmethod{} outperforms all previous works in the high-resource machine translation scenario ($>$ 4M sentence pairs).

\paragraph{WMT14 En-Fr} We further conduct experiments on the WMT14 En-Fr bilingual translation task. Table \ref{tab:wmt14-en-de} \mourmethod{} shows that \mourmethod{} still brings significant improvement to the downstream task with large-scale machine translation fine-tuning data ($>$ 40M sentence pairs).

\paragraph{WMT16 En-Ro} For the low-resource setting ($<$ 1M sentence pairs), there is an average gain of +4 BLEU points compared to the Transformer baseline in Table \ref{tab:wmt16-en-ro}. With the same back-translation data, \mourmethod{} further improves the model performance and still beats other baselines.

\paragraph{WMT-10} For the multilingual translation, we compare \mourmethod{} with the strong multilingual pre-trained models in Table \ref{tab:wmt10:x2e} and Table \ref{tab:wmt10:e2x}, such as mBART \cite{mbart}. It is notable our method outperforms large pre-trained model mBART with 1024 hidden size by a large margin (+1$\sim$2 BLEU points). Plus, there is a +1.5 BLEU gain over XLM-R, whose encoder and decoder are initialized by the cross-lingual pre-trained encoder \cite{xlmt}.

\paragraph{WebNLG}
Table \ref{tab:WebNLG} presents the performance on the data-to-text generation task, showing that \ourmethod{} outperforms multilingual sequence-to-sequence pre-training baselines mBART and mT5 by +2 ROUGE-L points on both languages.

\begin{table}[t]
\small
\begin{center}
\resizebox{1.0\columnwidth}{!}{
\begin{tabular}{l|cccc}
\toprule
\bf Model & \bf En$\rightarrow$De & \bf De$\rightarrow$En & \bf En$\rightarrow$Fr & \bf Fr$\rightarrow$En \\ 
\midrule
Transformer \cite{transformer} & 27.8 & 30.7 & 38.2 & 37.4 \\
\midrule
mBERT \cite{bert}  & 28.0  & 30.8  & 38.0 & 37.8 \\
XLM-R~\cite{xlmr} & 29.4  & 31.4  & 39.5  & 38.7  \\
mBART~\cite{xlmr} & 29.5  & 33.2  & 42.0  & 39.2  \\
mT5~\cite{xlmr}   & 28.8  & 32.1  & 39.8  & 38.6  \\
\bf \mourmethod{} (ours)      & \bf 30.6     & \bf 34.0 & \bf 42.9   & \bf 39.8 \\
\bottomrule
\end{tabular}}
\end{center}
\caption{Comparison with other pre-training approaches on the WMT14 En-De and WMT14 En-Fr benchmark.
}
\label{tab:wmt14-en-de}
\end{table}

\begin{table}[t]
\small
\begin{center}
\resizebox{1.0\columnwidth}{!}{
\begin{tabular}{l|ccc}
\toprule
\bf Model & \bf  En$\to$Ro & \bf Ro$\to$En & \bf \makecell[c]{Ro$\to$En (+BT)} \\ 
\midrule
Transformer \cite{transformer}  & 34.0 & 33.3 & 36.4  \\
\midrule
XLM \cite{xlm}         & -    & 35.6 & 38.5 \\
MASS \cite{mass}       & -    & -    & 39.1 \\
BART \cite{bart}       & -    & -    & 38.0 \\
BART-En \cite{mbart}   & 36.0 & 35.8 & 37.4  \\
BART-Ro \cite{mbart}   & 37.6 & 36.8 & 38.1 \\
XLM-R \cite{xlmr}      & 35.6 & 35.8 & -     \\
mBART \cite{mbart}     & 37.7 & 37.8 & 38.8  \\
mT5 \cite{mbart}       & 37.1 & 37.2 & 38.0  \\
\bf \mourmethod{} (ours)  & \bf 38.3  & \bf 38.0 &  \bf 39.3 \\
\bottomrule
\end{tabular}}
\end{center}
\caption{Comparison with other pre-training methods on the WMT16 En-Ro benchmark. }
\label{tab:wmt16-en-ro}
\end{table}

\begin{table*}[t]
\centering
\resizebox{1.0\textwidth}{!}{
\begin{tabular}{l|l|c|cccccccccc|c}
\toprule
\multicolumn{2}{l|}{\bf En$\rightarrow$X test sets} & \bf \#Params & \bf Fr & \bf Cs &\bf De & \bf Fi & \bf Lv & \bf Et & \bf Ro & \bf Hi & \bf Tr & \bf Gu & \bf Avg$_{\bm{10}}$ \\
\midrule
1$\rightarrow$1 & BiNMT \cite{transformer} &242M/10M& 36.3 & 22.3 & 40.2 & 15.2 & 16.5 & 15.0 & 23.0 & 12.2 & 13.3 & 7.9 & 20.2\\
\midrule
\multirow{5}{*}{1$\rightarrow$N} & MNMT \cite{transformer}  &242M& 34.2 & 20.9 & 40.0 & 15.0 & 18.1 & 20.9 & 26.0 & 14.5 & 17.3 & 13.2 & 22.0 \\
& mBART~\cite{mbart} &611M& 33.7 & 20.8 & 38.9 & 14.5 & 18.2 & 20.5 & 26.0 & 15.3 & 16.8 & 12.9 & 21.8 \\
& XLM-R \cite{xlmr} &362M& 34.7 &  21.5 &  40.1 &  15.2 & 18.6 &  20.8 &  26.4 & 15.6 &  17.4 & 14.9 & 22.5 \\
& \bf \ourmethod{} (ours)  & 430M &  \bf 36.0 & \bf 22.4 & \bf 42.1 & \bf 16.5 & \bf 19.7 &\bf  21.5 & \bf 27.0 &\bf  17.4 & \bf 18.6 & \bf 16.3 &\bf  23.8 \\
\midrule
\multirow{5}{*}{N$\rightarrow$N} & MNMT \cite{transformer} & 242M & 34.2 & 21.0 & 39.4 & 15.2 & 18.6 & 20.4 & 26.1 & 15.1 & 17.2 & 13.1 & 22.0 \\
& mBART~\cite{mbart} & 611M & 32.4 & 19.0 & 37.0 & 13.2 & 17.0 & 19.5 & 25.1 & 15.7 & 16.7 & 14.2 & 21.0 \\
& XLM-R \cite{xlmr} &362M& 34.2 &  21.4 &  39.7 &  15.3 & 18.9 &  20.6 &  26.5 & 15.6 &  17.5 & 14.5 & 22.4 \\
& \bf \mourmethod{} (ours) &430M & \bf 35.0 & \bf 21.8 & \bf 40.2 & \bf 16.1 & \bf 19.2 & \bf 21.9 & \bf 26.7 & \bf 16.2 & \bf 17.9 & \bf 14.4 & \bf 22.9 \\
\bottomrule
\end{tabular}}
\caption{En$\rightarrow$X evaluation results for bilingual (1$\rightarrow$1), one-to-many (1$\rightarrow$N), and many-to-many (N$\rightarrow$N) models on WMT-10. The languages are ordered from high-resource languages (left) to low-resource languages (right).}
\label{tab:wmt10:e2x}
\end{table*}

\begin{table*}[t]
\centering
\resizebox{1.0\textwidth}{!}{
\begin{tabular}{l|l|c|cccccccccc|c}
\toprule
\multicolumn{2}{l|}{\bf X$\rightarrow$En test sets} &\bf  \#Params & \bf Fr & \bf Cs & \bf De & \bf Fi & \bf Lv & \bf Et & \bf Ro & \bf Hi & \bf Tr & \bf Gu & \bf Avg$_{\bm{10}}$ \\
\midrule
1$\to$1 & BiNMT \cite{transformer} &242M/10M& 36.2 &  28.5 &  40.2 & 19.2 & 17.5 & 19.7 & 29.8 &  14.1 &  15.1 &  9.3 & 23.0\\
\midrule
\multirow{5}{*}{N$\to$1} & MNMT \cite{transformer}  &242M& 34.8 & 29.0 & 40.1 & 21.2 & 20.4 & 26.2 & 34.8 &  22.8 & 23.8 & 19.2 & 27.2 \\
& mBART~\cite{mbart} &611M& 36.2 & 29.9 &  40.0 &  22.2 &  20.6 & 27.2 & 37.2 &  23.3 &  25.7 &  21.7 & 28.4 \\
& XLM-R \cite{xlmr} &362M& 35.6 &  30.2 &  40.9 &  22.7 & 21.7 &  28.4 &  37.3 & 25.4 &  26.2 & 22.6 & 29.1 \\
& \bf \ourmethod{} (ours)  & 430M &  \bf 36.9 & \bf 31.8 & \bf 42.4 & \bf 23.2 & \bf 22.5 & \bf 29.4 & \bf 37.9 & \bf 27.2 & \bf 27.9 & \bf 22.9 & \bf 30.2 \\
\midrule
\multirow{5}{*}{N$\to$N} & MNMT \cite{transformer} & 242M & 35.9 & 29.2 &  40.0 & 21.1 & 20.4 & 26.3 & 35.5 & 23.6 &  24.3 & 20.6 & 27.7\\
& mBART~\cite{mbart} & 611M & 34.8 & 28.9 & 39.4 & 20.7 & 20.2 & 25.8 & 35.9 & 22.5 & 25.0 & 21.9 & 27.5 \\
& XLM-R \cite{xlmr} &362M& 35.7 & 30.3 & 41.0 & 22.2 & 21.3 & 28.1 & 37.0 & 25.4 & 26.1 & 21.9 & 28.9 \\
& \bf \mourmethod{} (ours) &430M& \bf 37.0 & \bf 31.1 & \bf 42.4 & \bf 22.7 & \bf 22.5 & \bf 28.1 & \bf 37.1 & \bf 25.3 & \bf 26.9 & \bf 22.7 & \bf 29.6 \\
\bottomrule
\end{tabular}}
\caption{X$\rightarrow$En evaluation results for bilingual (1$\rightarrow$1), one-to-many (1$\rightarrow$N), and many-to-many (N$\rightarrow$N) models on WMT-10. The languages are ordered from high-resource languages (left) to low-resource languages (right).}
\label{tab:wmt10:x2e}
\end{table*}

\begin{table}[htb]
\centering
\resizebox{1.0\columnwidth}{!}{
\small
\begin{tabular}{lccc}
\toprule
{\bf Model}  & {\bf En} & {\bf Ro} \\
&{ RG-1/RG-2/RG-L} & { RG-1/RG-2/RG-L} \\ 
\midrule
mBART   \cite{mbart}                   &  83.4/63.1/70.3 & 34.8/13.4/33.0 \\
mT5$_{\text{small}}$    \cite{gem}         & 78.8/59.2/67.2  & 29.7/10.5/28.4 \\
mT5$_{\text{base}}$    \cite{gem}         & 82.3/62.1/69.7  & 33.0/12.7/31.3 \\
\textbf{\mourmethod{} (ours)}   & \bf 83.8/63.9/71.2   & \bf 35.2/15.0/33.4 \\
\bottomrule
\end{tabular}
}
\caption{Results on data-to-text generation (WebNLG).}
\label{tab:WebNLG}
\end{table}

\section{Analysis}
\label{analysis}

\paragraph{Ablation Study}
To analyze the effect of the proposed pre-training and fine-tuning strategies, we conduct an ablation study of each component of our method in Table \ref{ablation_study}. Experiment {\large{\ding{175}}} and {\large{\ding{177}}} verify the merits of the \dtask{} and \gtask{}.
Furthermore, experiment {\large{\ding{178}}} shows that our model with the \gtask{} task obtains the best performance by jointly fine-tuning generator ($\mathcal{G}$) and discriminator ($\mathcal{D}$).

\begin{table}[htb]
\centering
\resizebox{1.0\columnwidth}{!}{
\begin{tabular}{c|l|cc|c}
\toprule
\bf ID & \bf Method          & $\bm{\mathcal{D}}$ & $\bm{\mathcal{G}}$ & \makecell[c]{\textbf{Xsum} \\ RG-1/RG-2/RG-L}   \\
\midrule
{\large{\ding{172}}} & Transformer w/o Pre-training  &  & \checkmark  & 32.36/11.46/25.47  \\ 
{\large{\ding{173}}} & {\large{\ding{172}}} + S2S-MLM & & \checkmark   & 44.44/21.25/36.22   \\
{\large{\ding{174}}} & {\large{\ding{173}}} + \Dtask{} &\checkmark &    & 42.11/18.58/33.21    \\
{\large{\ding{175}}} & {\large{\ding{173}}} + \Dtask{} &  &\checkmark   & 44.28/21.14/36.24    \\
\midrule
{\large{\ding{176}}} & {\large{\ding{175}}} +  \Gtask{} &\checkmark&  &    42.41/18.98/34.31   \\
{\large{\ding{177}}} & {\large{\ding{175}}} +  \Gtask{} &  &\checkmark &    44.74/21.47/36.40  \\     
{\large{\ding{178}}} & {\large{\ding{175}}} +  \Gtask{} &\checkmark&\checkmark& \bf 45.36/21.98/36.84   \\
\bottomrule
\end{tabular}
}
\caption{Ablation study of our method on the test set of the abstractive summarization benchmark XSum, where \ourmethod{} is fine-tuned on the downstream task with different pre-training and fine-tuning strategies.}
\label{ablation_study}
\vspace{-10pt}
\end{table}

\paragraph{Low-resource Setting}
\begin{figure}[t]
    \centering
    \subfigure[En$\to$Ro]{
    \includegraphics[width=0.45\columnwidth]{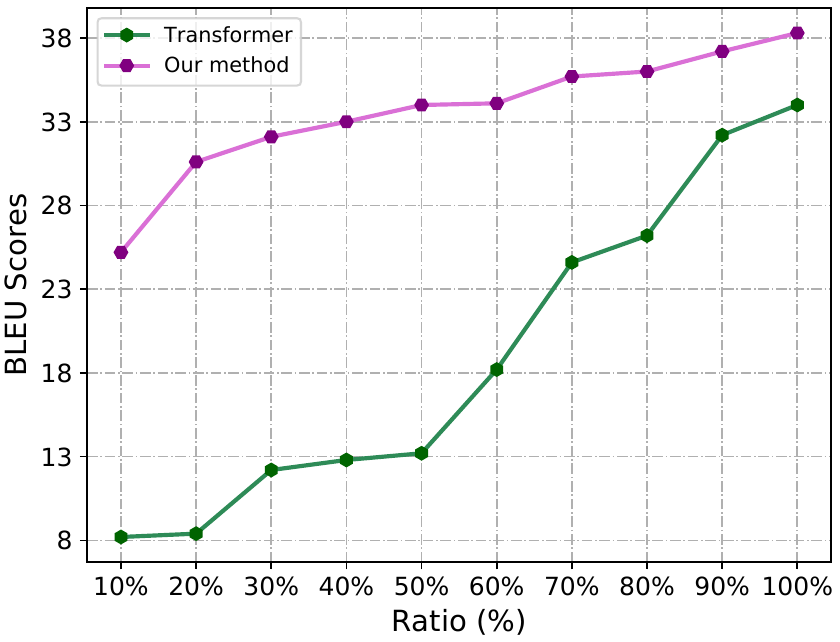}
    \label{low_resource_en2ro}
    }
    \subfigure[Ro$\to$En]{
    \includegraphics[width=0.45\columnwidth]{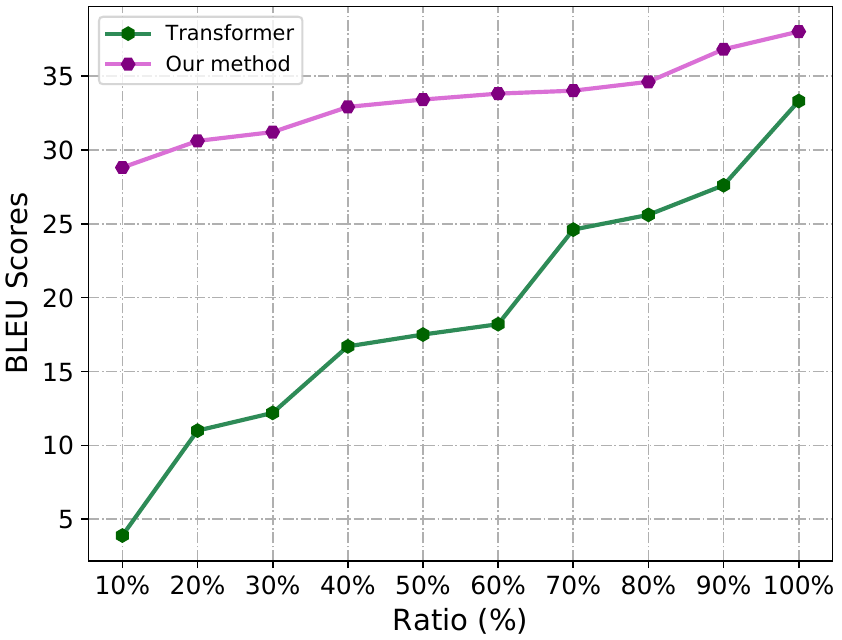}
    \label{low_resource_ro2en}
    }
    \caption{Comparison between Transformer and our method on WMT-16 (a) En$\to$Ro and (b) Ro$\to$ En.} 
    \label{low_resource}
\end{figure}
To further analyze the performance of \mourmethod{} given different sizes of downstream parallel data, we randomly extract $K$ percentage of the whole sentence pairs as the fine-tuned parallel data from the full WMT-16 En$\to$Ro training data. We set $K$ = \{$10\%$, $20\%$, \dots, $100\%$\} and compare our method with the Transformer baseline model. Figure \ref{low_resource} shows the BLEU points of our pre-trained multilingual model and the baseline. When the parallel data size is small, the baseline without pre-trained model produces unsatisfactory results. Similarly, in Figure \ref{low_resource_en2ro}, \ourmethod{} fine-tuned on nearly half data (purple line, 50\%) defeats the baseline trained on all pairs (green line, 100\%), exemplifying the effectiveness of our method in low-resource scenarios.

\paragraph{Discussion on Discriminator}
\begin{figure}[t]
    \centering
    \subfigure[Discriminator Weight]{
    \includegraphics[width=0.45\columnwidth]{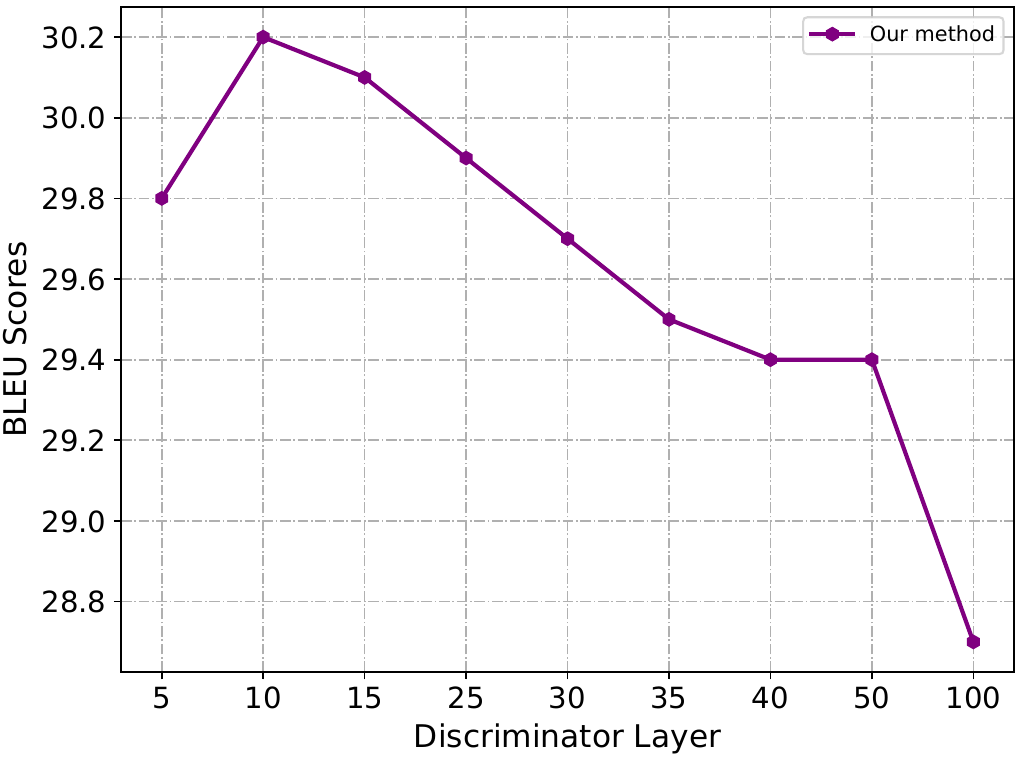}
    \label{discriminator_weight}
    }
    \subfigure[Discriminator Layer]{
    \includegraphics[width=0.45\columnwidth]{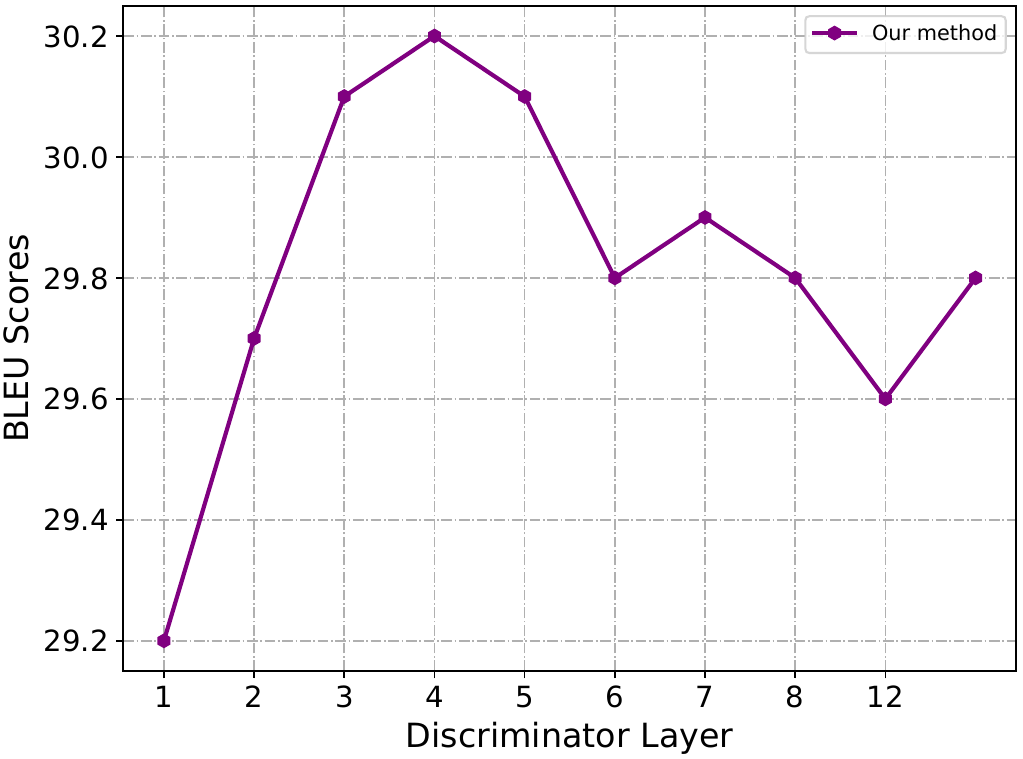}
    \label{discriminator_layer}
    }
    \caption{Effect of (a) discriminator weight and (b) Discriminator layer on the WMT14 En$\to$De task.} 
    \label{discriminator}
\end{figure}
The weight value $\lambda$ and layer number of the discriminator are key factors to our pre-training task. As shown in Figure \ref{discriminator}, we vary discriminator weight in Figure \ref{discriminator_weight} to find a balance between the generator and discriminator objective. To this end, we study the performance of \ourmethod{} with different $\lambda$, where $\lambda$ ranges from $5.0$ to $100.0$. When the weight of the discriminator is 10.0, multiple pre-training tasks are balanced. Moreover, we find it more efficient to have a tiny discriminator (3 $\sim$ 6 layers) in Figure \ref{discriminator_layer}.

\begin{figure}[t]
    \centering
    \subfigure[$11$-th]{
    \includegraphics[width=0.21\columnwidth]{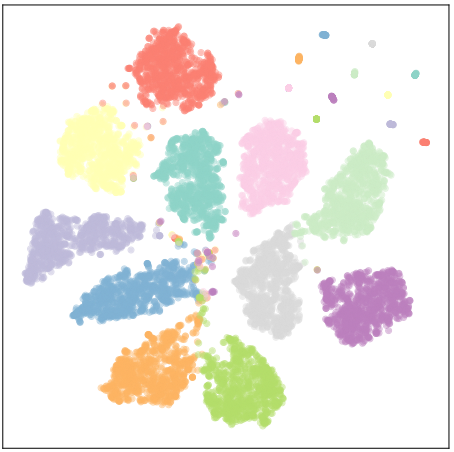}
    \label{tsne_1}
    }
    \subfigure[$12$-th]{
    \includegraphics[width=0.21\columnwidth]{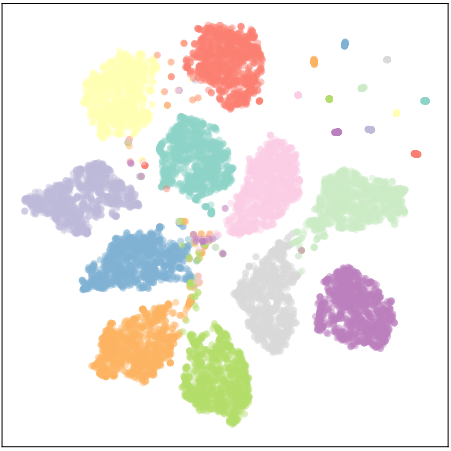}
    \label{tsne_2}
    }
    \subfigure[$11$-th]{
    \includegraphics[width=0.21\columnwidth]{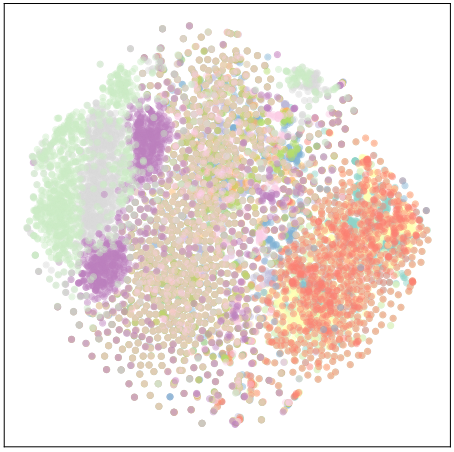}
    \label{tsne_3}
    }
    \subfigure[$12$-th]{
    \includegraphics[width=0.21\columnwidth]{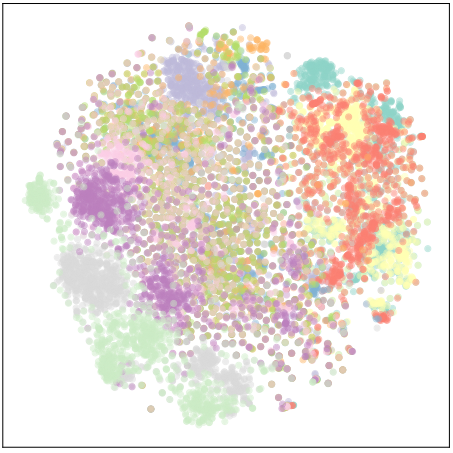}
    \label{tsne_4}
    }
    \caption{(a) and (b) are representations of the baseline from the $11$-th and $12$-th encoder layers while (c) and (d) are counterparts of the fine-tuned model. Each color denotes one language (11 languages in WMT-10).} 
    \label{tsne_figures}
\end{figure}

\paragraph{Multilingual Representations}
We randomly select 1000 parallel sentences of each language in WMT-10 and visualize their representations \cite{t_SNE} of the last two encoder layers in Figure \ref{tsne_figures} using our multilingual model fine-tuned on WMT-10 and the multilingual baseline. The first hidden state of the encoder is adopted as the sentence representation. Compared to Figure \ref{tsne_1} and \ref{tsne_2} of the baseline, different languages become closer and likely to overlap with each other in Figure \ref{tsne_3} and \ref{tsne_4} of our method, demonstrating that our method effectively aligns representations of different languages to the shared space.

\begin{table}[t]
\centering
\resizebox{1.0\columnwidth}{!}{
\begin{tabular}{l|c|ccc}
\toprule
\bf Model                     & \bf \#Params &  \bf Avg$_{\bm{X \to En}}$  & \bf Avg$_{\bm{En \to Y}}$ & \bf Avg$_{\bm{X \to Y}}$   \\
\midrule
M2M-124$_{\text{base}}$  \cite{flores}   &  175M       & 15.43 & 12.02 &   5.85       \\
M2M-124$_{\text{large}}$ \cite{flores}   &  615M       & 20.03 & 16.21  &  7.66          \\
DeltaLM + Zcode \cite{microsoft_wmt2021}     &  711M      & 30.39 &  23.52 &  11.21          \\    
\bf \mourmethod{} (ours)    & 430M  & \bf 30.70    & \bf 24.83  &  \bf 13.65         \\
\bottomrule
\end{tabular}
}
\caption{Massively multilingual translation average results ($102 \times 101$ translation directions) on the devtest sets of the flores benchmark.}
\label{flores}
\end{table}

\paragraph{Massively Multilingual Translation}
We compare \mourmethod{} with the state-of-the-art multilingual NMT model M2M-124 \cite{flores}. M2M-124$_{\text{large}}$ and DeltaLM + Zcode both have a large hidden size of 1024. Our pre-trained model is fine-tuned on the same training data as DeltaLM + Zcode \cite{microsoft_wmt2021}. Compared to M2M-124$_{\text{large}}$, \mourmethod{} with fewer training data and only 430M parameters depends more on the transferability of the cross-lingual pre-training model. In Table \ref{flores}, our method outperforms the DeltaLM + Zcode in zero-shot translation direction (Avg$_{X \to Y}$) by +1.5 BLEU points, benefiting from our pre-trained model in cross-lingual zero-shot transfer.

\paragraph{Comparison of Pre-training Cost}
Our English pre-trained model \ourmethod{} is trained for nearly 2 weeks on 128 A100 GPUs (40GB), with 500K training steps and a batch size of 8K sequences. Compared to the re-implemented T5 \cite{t5}, our method is only 0.5 times slower than T5 with the same training steps but gets a significant improvement on the machine translation, text summarization, and data-to-text generation tasks.
\begin{table}[t]
\centering
\resizebox{1.0\columnwidth}{!}{
\begin{tabular}{lcccccccc}
\toprule
\bf Model     & \bf MNLI & \bf SST-2 & \bf MRPC & \bf RTE & \bf QNLI  & \bf QQP & \bf Avg$_{\bm{6}}$ \\ \midrule
BERT \cite{bert}                   &  84.5       &  93.2      &  87.3     &  68.6      &  91.7             &  91.3      & 86.1    \\
XLNet  \cite{xlnet}                &  86.8       &  94.7      &  88.2     &  74.0      &  91.7             &  91.4      & 87.8\\
RoBERTa  \cite{roberta}            &  87.6       &  94.8      &  90.2     &  78.7      &  92.8             &  91.9      & 89.3  \\
\mourmethod{} ($\mathcal{D}$)       &  89.0       &  94.7      & \bf 90.6  & 83.2       & 93.9              &  91.7      & 90.5\\
\mourmethod{} ($\mathcal{G}$)       &  \bf 89.3   & \bf 95.0   & 90.5      & \bf 85.0   & \bf 94.2          & \bf 92.0    & \bf 91.0\\
\bottomrule
\end{tabular}}
\caption{Results of base-setting models on the valid set of GLUE. We report accuracy for classification tasks.}
\label{tab:glue}
\end{table}

\begin{table}[t]
\centering
\resizebox{1.0\columnwidth}{!}{
\begin{tabular}{lcccccc}
\toprule
\bf Models & \bf En &\bf  De & \bf Th &\bf Tr &  \bf Vi & \bf Avg$_{\bm{15}}$ \\
\midrule
\multicolumn{7}{l}{\textit{\textit{Fine-tuning on English training set (Cross-lingual zero-shot transfer)}}} \\
\midrule
XLM \cite{xlm}            & 85.0      & 77.8 & 73.2  &72.5  & 76.1  & 75.1 \\
mT5 \cite{mt5}            & 84.7      & 77.4 & 73.2  &72.8  & 74.2  & 75.4 \\
\mourmethod{} ($\mathcal{D}$)  & 85.0      & 78.6 & \bf 74.3  & \bf 74.4  & \bf 77.2  & \bf 75.8 \\
\mourmethod{} ($\mathcal{G}$)  & \bf 86.3  & \bf 79.0 & 74.2  &74.5  & 76.5  & 75.5 \\
\midrule
\multicolumn{7}{l}{\textit{Fine-tuning on each training set (Translate-train)}} \\
\midrule
XLM \cite{xlm}            & 85.0 & 80.3  &75.5  &74.7  &76.6  & 76.7 \\
mT5 \cite{mt5}            & 84.7 & -  &-  &-  &-  & - \\
\mourmethod{} ($\mathcal{D}$)  & 85.0 & 80.7  & 76.9 &74.4  &79.1  & 77.9 \\
\mourmethod{} ($\mathcal{G}$))  & \bf 86.3 & \bf 80.8  &\bf 77,4  &\bf 74.5  &\bf 79.2  & \bf 78.0 \\
\midrule
\multicolumn{7}{l}{\textit{Fine-tuning on all training sets (Translate-train-all)}} \\
\midrule
XLM \cite{xlm}            &85.0 &80.3 &76.0 &75.6 &78.5 &77.8 \\
mT5 \cite{mt5}            &82.0 &77.7 &75.0 &74.8 &74.5 &75.9 \\
\mourmethod{} ($\mathcal{D}$)  &\bf 87.3 & \bf 83.1 & \bf 80.3 & \bf 79.9 &  81.3 &   80.5\\
\mourmethod{} ($\mathcal{G}$)  &87.2 & 82.7 & 79.8 & 79.6  & \bf 81.6  & \bf  80.6 \\
\bottomrule
\end{tabular}}
\caption{Analysis of multilingual classification on the XNLI test set. The evaluation metric is accuracy (\%).}
\label{xnli}
\end{table}

\begin{figure}[t]
\begin{center}
	\includegraphics[width=0.6\columnwidth]{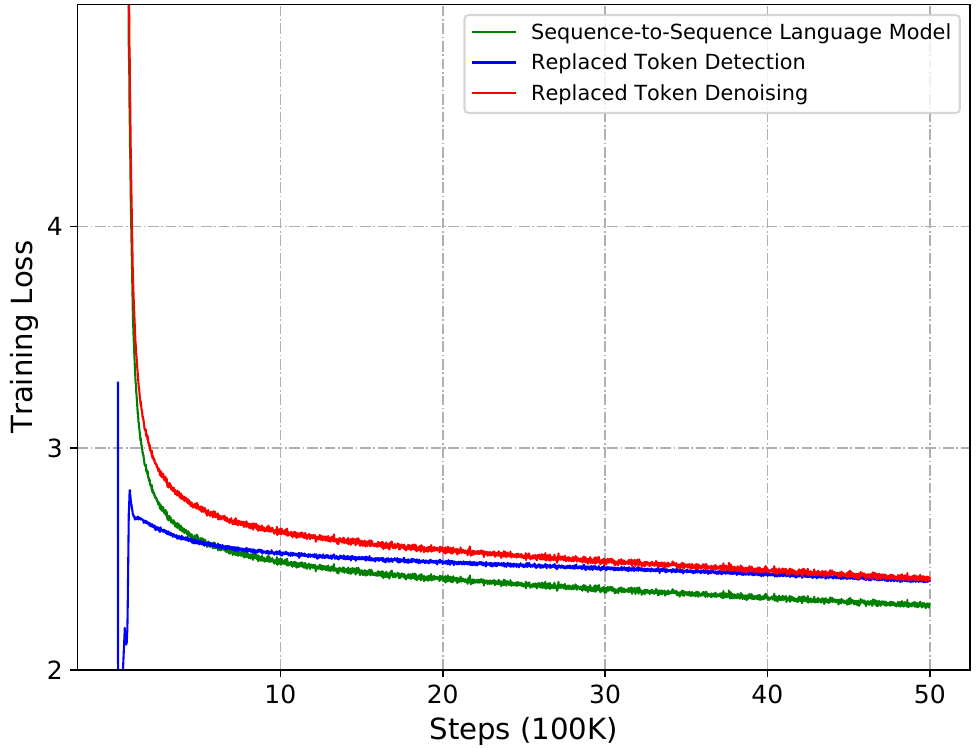}
	\caption{The training loss of sequence-to-sequence language modeling, \dtask{}, and \gtask{} in the pre-training stage of our English pre-trained model \ourmethod{}.}
	\label{training_loss}
\end{center}
\end{figure}

\begin{figure}[t]
\begin{center}
	\includegraphics[width=0.6\columnwidth]{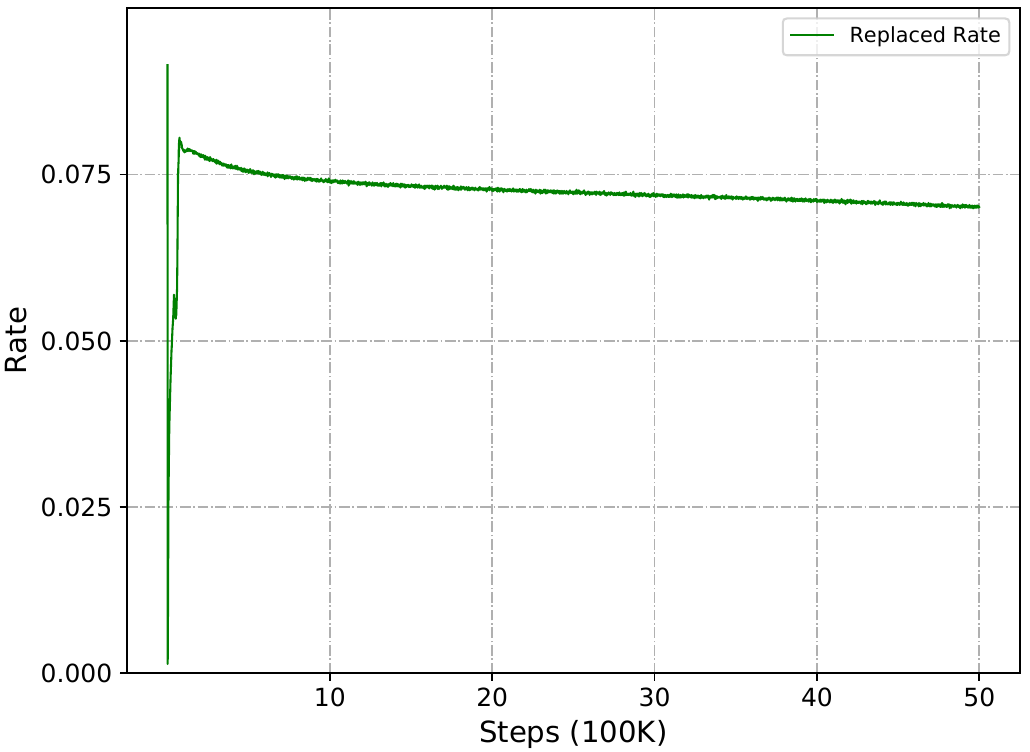}
	\caption{The replaced rate of the \gtask{} task in the pre-training stage of our English pre-trained model \ourmethod{}}.
	\label{replaced_rate}
\end{center}
\end{figure}

\paragraph{Training of \gtask{}}
To fully understand the training procedure of the \gtask{}, we plot the training loss of sequence-to-sequence masked language modeling $L_{\mathcal{G}}$, \dtask{}, and \gtask{} in Figure \ref{training_loss}. Furthermore, we investigate how many tokens in the target sentence are replaced with the misclassified tokens by discriminator in Figure \ref{replaced_rate}. We define $p_{r}$ as the replaced rate in the target gold sentence. Nearly 7.5\% tokens of the target previous tokens are replaced with the misclassified tokens to construct the noisy input samples for the generator decoder.

\paragraph{Language Understanding} 
Our method can be easily extended to various downstream language understanding tasks. We use the GLUE benchmark \cite{glue} to estimate English pre-trained model \ourmethod{} and the XNLI dataset \cite{xnli} to evaluate the capability of the multilingual language understanding. Our method is tested on each language separately by fine-tuning generator ($\mathcal{G}$) or discriminator ($\mathcal{D}$) on the XNLI dataset. In Table \ref{tab:glue}, Our English pre-trained model performs better than RoBERTa. Additionally, our pre-trained model outperforms the previous cross-lingual pre-trained encoder XLM and pre-trained encoder-decoder model mT5 in Table \ref{xnli}.

\section{Related Work}
\paragraph{Pre-training for Generation} Language modeling based on the self-supervised learning training objective and large-scale data has been widely used to acquire contextual representations. Pre-training a large Transformer encoder \cite{transformer,bert,spanbert,roberta} with the masked language modeling (MLM) task brings significant improvement for various downstream natural language understanding (NLU) tasks. Many enhanced versions of MLM tasks \cite{spanbert,ernie:baidu,roberta,electra} are proposed to further enhance the capability of the pre-trained model. Besides, pre-training a Transformer decoder \cite{gpt,gpt2,gpt3} is beneficial for unconditional text generation.
There have been numerous attempts for pre-training a sequence-to-sequence Transformer model by adding generative training objectives, such as MASS \cite{mass} and BART \cite{bart}.  Furthermore, T5 \cite{t5} explores different pre-training tasks and proposes to corrupt consecutive span of tokens for pre-training. Different from previous works, our work focuses on leveraging the auxiliary discriminator ameliorate encoder-decoder pre-training on language generation tasks.

\paragraph{Multilingual Pre-training} Inspired the success of pre-training in a single language such as English, recent works \cite{xlm,xlmr,crop,alm,infoxlm,hlt_mt,um4,microsoft_wmt2021} aim to learn cross-lingual representations with different training objectives in multiple languages. For the sequence-to-sequence model, mBART \cite{mbart} pre-trains a Transformer model by denoising training objective in multiple languages. mT5 \cite{mt5} extends the span corruption task for multilingual training and mT6 \cite{mt6} amplify generation task by introducing a partially non-autoregressive objective. Along the line of research, different multilingual pre-trained models \cite{xlmt,xnlg} are proposed to solve downstream cross-lingual generation tasks.

\section{Conclusion}
In this work, we introduce \ourmethod{}, a state-of-the-art pre-training encoder-decoder framework for both language generation and understanding tasks trained on large-scale corpora. Our GAN-style models are pre-trained with \dtask{} and \gtask{} by introducing an auxiliary discriminator.
Extensive experiments prove the effectiveness of $\ourmethod{}$ on various language generation and translation benchmark datasets. We further verify the capability of the pre-trained model on multiple downstream understanding tasks.

\section*{Acknowledgments}
This work was supported in part by the National Natural Science Foundation of China (Grant Nos. 62276017, U1636211, 61672081), the 2022 Tencent Big Travel Rhino-Bird Special Research Program, and the Fund of the State Key Laboratory of Software Development Environment (Grant No. SKLSDE-2021ZX-18). 

\bibliography{custom}

\begin{thebibliography}{53}
\expandafter\ifx\csname natexlab\endcsname\relax\def\natexlab#1{#1}\fi

\bibitem[{Agic and Vulic(2019)}]{jw300}
Zeljko Agic and Ivan Vulic. 2019.
\newblock {JW300:} {A} wide-coverage parallel corpus for low-resource
  languages.
\newblock In \emph{ACL 2019}, pages 3204--3210.

\bibitem[{Bao et~al.(2021)Bao, Dong, Wang, Yang, and Wei}]{s2s_ft}
Hangbo Bao, Li~Dong, Wenhui Wang, Nan Yang, and Furu Wei. 2021.
\newblock s2s-ft: Fine-tuning pretrained transformer encoders for
  sequence-to-sequence learning.
\newblock \emph{CoRR}, abs/2110.13640.

\bibitem[{Bao et~al.(2020)Bao, Dong, Wei, Wang, Yang, Liu, Wang, Gao, Piao,
  Zhou, and Hon}]{unilmv2}
Hangbo Bao, Li~Dong, Furu Wei, Wenhui Wang, Nan Yang, Xiaodong Liu, Yu~Wang,
  Jianfeng Gao, Songhao Piao, Ming Zhou, and Hsiao{-}Wuen Hon. 2020.
\newblock Unilmv2: Pseudo-masked language models for unified language model
  pre-training.
\newblock In \emph{ICML 2020}, volume 119, pages 642--652.

\bibitem[{Chi et~al.(2021{\natexlab{a}})Chi, Dong, Ma, Huang, Singhal, Mao,
  Huang, Song, and Wei}]{mt6}
Zewen Chi, Li~Dong, Shuming Ma, Shaohan Huang, Saksham Singhal, Xian{-}Ling
  Mao, Heyan Huang, Xia Song, and Furu Wei. 2021{\natexlab{a}}.
\newblock {mT6}: Multilingual pretrained text-to-text transformer with
  translation pairs.
\newblock In \emph{EMNLP 2021}, pages 1671--1683.

\bibitem[{Chi et~al.(2020)Chi, Dong, Wei, Wang, Mao, and Huang}]{xnlg}
Zewen Chi, Li~Dong, Furu Wei, Wenhui Wang, Xian{-}Ling Mao, and Heyan Huang.
  2020.
\newblock Cross-lingual natural language generation via pre-training.
\newblock In \emph{AAAI 2020}, pages 7570--7577.

\bibitem[{Chi et~al.(2021{\natexlab{b}})Chi, Dong, Wei, Yang, Singhal, Wang,
  Song, Mao, Huang, and Zhou}]{infoxlm}
Zewen Chi, Li~Dong, Furu Wei, Nan Yang, Saksham Singhal, Wenhui Wang, Xia Song,
  Xian{-}Ling Mao, Heyan Huang, and Ming Zhou. 2021{\natexlab{b}}.
\newblock Infoxlm: An information-theoretic framework for cross-lingual
  language model pre-training.
\newblock In \emph{NAACL 2021}, pages 3576--3588.

\bibitem[{Clark et~al.(2020)Clark, Luong, Le, and Manning}]{electra}
Kevin Clark, Minh-Thang Luong, Quoc~V. Le, and Christopher~D. Manning. 2020.
\newblock {ELECTRA}: Pre-training text encoders as discriminators rather than
  generators.
\newblock In \emph{ICLR}.

\bibitem[{Conneau et~al.(2020)Conneau, Khandelwal, Goyal, Chaudhary, Wenzek,
  Guzm{\'{a}}n, Grave, Ott, Zettlemoyer, and Stoyanov}]{xlmr}
Alexis Conneau, Kartikay Khandelwal, Naman Goyal, Vishrav Chaudhary, Guillaume
  Wenzek, Francisco Guzm{\'{a}}n, Edouard Grave, Myle Ott, Luke Zettlemoyer,
  and Veselin Stoyanov. 2020.
\newblock Unsupervised cross-lingual representation learning at scale.
\newblock In \emph{ACL 2020}, pages 8440--8451.

\bibitem[{Conneau and Lample(2019)}]{xlm}
Alexis Conneau and Guillaume Lample. 2019.
\newblock Cross-lingual language model pretraining.
\newblock In \emph{NeurIPS 2019}, pages 7057--7067.

\bibitem[{Conneau et~al.(2018)Conneau, Lample, Rinott, Williams, Bowman,
  Schwenk, and Stoyanov}]{xnli}
Alexis Conneau, Guillaume Lample, Ruty Rinott, Adina Williams, Samuel~R Bowman,
  Holger Schwenk, and Veselin Stoyanov. 2018.
\newblock Xnli: Evaluating cross-lingual sentence representations.
\newblock \emph{arXiv preprint arXiv:1809.05053}.

\bibitem[{Devlin et~al.(2019)Devlin, Chang, Lee, and Toutanova}]{bert}
Jacob Devlin, Ming{-}Wei Chang, Kenton Lee, and Kristina Toutanova. 2019.
\newblock {BERT:} {P}re-training of deep bidirectional transformers for
  language understanding.
\newblock In \emph{NAACL 2019}, pages 4171--4186.

\bibitem[{Dong et~al.(2019)Dong, Yang, Wang, Wei, Liu, Wang, Gao, Zhou, and
  Hon}]{unilm}
Li~Dong, Nan Yang, Wenhui Wang, Furu Wei, Xiaodong Liu, Yu~Wang, Jianfeng Gao,
  Ming Zhou, and Hsiao{-}Wuen Hon. 2019.
\newblock Unified language model pre-training for natural language
  understanding and generation.
\newblock In \emph{NeurIPS 2019}, pages 13042--13054.

\bibitem[{El{-}Kishky et~al.(2020)El{-}Kishky, Chaudhary, Guzm{\'{a}}n, and
  Koehn}]{ccalign}
Ahmed El{-}Kishky, Vishrav Chaudhary, Francisco Guzm{\'{a}}n, and Philipp
  Koehn. 2020.
\newblock Ccaligned: {A} massive collection of cross-lingual web-document
  pairs.
\newblock In \emph{EMNLP 2020}, pages 5960--5969.

\bibitem[{Fedus et~al.(2018)Fedus, Goodfellow, and Dai}]{maskgan}
William Fedus, Ian~J. Goodfellow, and Andrew~M. Dai. 2018.
\newblock Maskgan: Better text generation via filling in the
  {\_}{\_}{\_}{\_}{\_}{\_}{\_}.
\newblock In \emph{ICLR 2018}.

\bibitem[{Gardent et~al.(2017)Gardent, Shimorina, Narayan, and
  Perez{-}Beltrachini}]{WebNLG}
Claire Gardent, Anastasia Shimorina, Shashi Narayan, and Laura
  Perez{-}Beltrachini. 2017.
\newblock The webnlg challenge: Generating text from {RDF} data.
\newblock In \emph{{INLG} 2017}, pages 124--133.

\bibitem[{Gehrmann et~al.(2021)Gehrmann, Adewumi, Aggarwal, Ammanamanchi,
  Anuoluwapo, Bosselut, Chandu, Clinciu, Das, Dhole, Du, Durmus, Dusek, Emezue,
  Gangal, Garbacea, Hashimoto, Hou, Jernite, Jhamtani, Ji, Jolly, Kumar,
  Ladhak, Madaan, Maddela, Mahajan, Mahamood, Majumder, Martins,
  McMillan{-}Major, Mille, van Miltenburg, Nadeem, Narayan, Nikolaev,
  Niyongabo, Osei, Parikh, Perez{-}Beltrachini, Rao, Raunak, Rodriguez,
  Santhanam, Sedoc, Sellam, Shaikh, Shimorina, Cabezudo, Strobelt, Subramani,
  Xu, Yang, Yerukola, and Zhou}]{gem}
Sebastian Gehrmann, Tosin~P. Adewumi, Karmanya Aggarwal, Pawan~Sasanka
  Ammanamanchi, Aremu Anuoluwapo, Antoine Bosselut, Khyathi~Raghavi Chandu,
  Miruna{-}Adriana Clinciu, Dipanjan Das, Kaustubh~D. Dhole, Wanyu Du, Esin
  Durmus, Ondrej Dusek, Chris Emezue, Varun Gangal, Cristina Garbacea,
  Tatsunori Hashimoto, Yufang Hou, Yacine Jernite, Harsh Jhamtani, Yangfeng Ji,
  Shailza Jolly, Dhruv Kumar, Faisal Ladhak, Aman Madaan, Mounica Maddela,
  Khyati Mahajan, Saad Mahamood, Bodhisattwa~Prasad Majumder, Pedro~Henrique
  Martins, Angelina McMillan{-}Major, Simon Mille, Emiel van Miltenburg, Moin
  Nadeem, Shashi Narayan, Vitaly Nikolaev, Rubungo~Andre Niyongabo, Salomey
  Osei, Ankur~P. Parikh, Laura Perez{-}Beltrachini, Niranjan~Ramesh Rao, Vikas
  Raunak, Juan~Diego Rodriguez, Sashank Santhanam, Jo{\~{a}}o Sedoc, Thibault
  Sellam, Samira Shaikh, Anastasia Shimorina, Marco Antonio~Sobrevilla
  Cabezudo, Hendrik Strobelt, Nishant Subramani, Wei Xu, Diyi Yang, Akhila
  Yerukola, and Jiawei Zhou. 2021.
\newblock The {GEM} benchmark: Natural language generation, its evaluation and
  metrics.
\newblock \emph{CoRR}, abs/2102.01672.

\bibitem[{Goodfellow et~al.(2014)Goodfellow, Pouget{-}Abadie, Mirza, Xu,
  Warde{-}Farley, Ozair, Courville, and Bengio}]{gan}
Ian~J. Goodfellow, Jean Pouget{-}Abadie, Mehdi Mirza, Bing Xu, David
  Warde{-}Farley, Sherjil Ozair, Aaron~C. Courville, and Yoshua Bengio. 2014.
\newblock Generative adversarial networks.
\newblock \emph{CoRR}, abs/1406.2661.

\bibitem[{Goyal et~al.(2021)Goyal, Gao, Chaudhary, Chen, Wenzek, Ju, Krishnan,
  Ranzato, Guzm{\'{a}}n, and Fan}]{flores}
Naman Goyal, Cynthia Gao, Vishrav Chaudhary, Peng{-}Jen Chen, Guillaume Wenzek,
  Da~Ju, Sanjana Krishnan, Marc'Aurelio Ranzato, Francisco Guzm{\'{a}}n, and
  Angela Fan. 2021.
\newblock The {FLORES-101} evaluation benchmark for low-resource and
  multilingual machine translation.
\newblock \emph{CoRR}, abs/2106.03193.

\bibitem[{Joshi et~al.(2019)Joshi, Chen, Liu, Weld, Zettlemoyer, and
  Levy}]{spanbert}
Mandar Joshi, Danqi Chen, Yinhan Liu, Daniel~S Weld, Luke Zettlemoyer, and Omer
  Levy. 2019.
\newblock Spanbert: Improving pre-training by representing and predicting
  spans.
\newblock \emph{arXiv preprint arXiv:1907.10529}.

\bibitem[{Kingma and Ba(2015)}]{adam}
Diederik~P. Kingma and Jimmy Ba. 2015.
\newblock Adam: {A} method for stochastic optimization.
\newblock In \emph{ICLR 2015}.

\bibitem[{Kudo and Richardson(2018)}]{sentencepiece}
Taku Kudo and John Richardson. 2018.
\newblock Sentencepiece: {A} simple and language independent subword tokenizer
  and detokenizer for neural text processing.
\newblock In \emph{EMNLP 2018}, pages 66--71.

\bibitem[{Ladhak et~al.(2020)Ladhak, Durmus, Cardie, and McKeown}]{wikilingua}
Faisal Ladhak, Esin Durmus, Claire Cardie, and Kathleen~R. McKeown. 2020.
\newblock Wikilingua: {A} new benchmark dataset for cross-lingual abstractive
  summarization.
\newblock \emph{CoRR}, abs/2010.03093.

\bibitem[{Lewis et~al.(2020)Lewis, Liu, Goyal, Ghazvininejad, Mohamed, Levy,
  Stoyanov, and Zettlemoyer}]{bart}
Mike Lewis, Yinhan Liu, Naman Goyal, Marjan Ghazvininejad, Abdelrahman Mohamed,
  Omer Levy, Veselin Stoyanov, and Luke Zettlemoyer. 2020.
\newblock {BART:} denoising sequence-to-sequence pre-training for natural
  language generation, translation, and comprehension.
\newblock In \emph{ACL 2020}, pages 7871--7880.

\bibitem[{Lin(2004)}]{rouge}
Chin-Yew Lin. 2004.
\newblock {ROUGE}: A package for automatic evaluation of summaries.
\newblock In \emph{ACL 2004}, pages 74--81.

\bibitem[{Liu(2019)}]{bertsum}
Yang Liu. 2019.
\newblock Fine-tune {BERT} for extractive summarization.
\newblock \emph{CoRR}, abs/1903.10318.

\bibitem[{Liu et~al.(2020)Liu, Gu, Goyal, Li, Edunov, Ghazvininejad, Lewis, and
  Zettlemoyer}]{mbart}
Yinhan Liu, Jiatao Gu, Naman Goyal, Xian Li, Sergey Edunov, Marjan
  Ghazvininejad, Mike Lewis, and Luke Zettlemoyer. 2020.
\newblock Multilingual denoising pre-training for neural machine translation.
\newblock \emph{TACL}, 8:726--742.

\bibitem[{Liu et~al.(2019)Liu, Ott, Goyal, Du, Joshi, Chen, Levy, Lewis,
  Zettlemoyer, and Stoyanov}]{roberta}
Yinhan Liu, Myle Ott, Naman Goyal, Jingfei Du, Mandar Joshi, Danqi Chen, Omer
  Levy, Mike Lewis, Luke Zettlemoyer, and Veselin Stoyanov. 2019.
\newblock Roberta: {A} robustly optimized {BERT} pretraining approach.
\newblock \emph{CoRR}, abs/1907.11692.

\bibitem[{Ma et~al.(2021)Ma, Dong, Huang, Zhang, Muzio, Singhal, Awadalla,
  Song, and Wei}]{deltalm}
Shuming Ma, Li~Dong, Shaohan Huang, Dongdong Zhang, Alexandre Muzio, Saksham
  Singhal, Hany~Hassan Awadalla, Xia Song, and Furu Wei. 2021.
\newblock Deltalm: Encoder-decoder pre-training for language generation and
  translation by augmenting pretrained multilingual encoders.
\newblock \emph{CoRR}, abs/2106.13736.

\bibitem[{Ma et~al.(2020)Ma, Yang, Huang, Chi, Dong, Zhang, Awadalla, Muzio,
  Eriguchi, Singhal, Song, Menezes, and Wei}]{xlmt}
Shuming Ma, Jian Yang, Haoyang Huang, Zewen Chi, Li~Dong, Dongdong Zhang,
  Hany~Hassan Awadalla, Alexandre Muzio, Akiko Eriguchi, Saksham Singhal, Xia
  Song, Arul Menezes, and Furu Wei. 2020.
\newblock {XLM-T:} scaling up multilingual machine translation with pretrained
  cross-lingual transformer encoders.
\newblock \emph{CoRR}, abs/2012.15547.

\bibitem[{Maaten and Hinton(2008)}]{t_SNE}
Laurens van~der Maaten and Geoffrey Hinton. 2008.
\newblock Visualizing data using t-sne.
\newblock \emph{JMLR}, 9(Nov):2579--2605.

\bibitem[{Narayan et~al.(2018)Narayan, Cohen, and Lapata}]{xsum}
Shashi Narayan, Shay~B. Cohen, and Mirella Lapata. 2018.
\newblock Don't give me the details, just the summary! topic-aware
  convolutional neural networks for extreme summarization.
\newblock In \emph{EMNLP 2018}, pages 1797--1807.

\bibitem[{Papineni et~al.(2002)Papineni, Roukos, Ward, and Zhu}]{bleu}
Kishore Papineni, Salim Roukos, Todd Ward, and Wei-Jing Zhu. 2002.
\newblock {BLEU}: {A} method for automatic evaluation of machine translation.
\newblock In \emph{ACL 2002}.

\bibitem[{Radford et~al.(2018)Radford, Narasimhan, Salimans, Sutskever
  et~al.}]{gpt}
Alec Radford, Karthik Narasimhan, Tim Salimans, Ilya Sutskever, et~al. 2018.
\newblock Improving language understanding by generative pre-training.

\bibitem[{Radford et~al.(2019)Radford, Wu, Child, Luan, Amodei, Sutskever
  et~al.}]{gpt2}
Alec Radford, Jeffrey Wu, Rewon Child, David Luan, Dario Amodei, Ilya
  Sutskever, et~al. 2019.
\newblock Language models are unsupervised multitask learners.

\bibitem[{Raffel et~al.(2020)Raffel, Shazeer, Roberts, Lee, Narang, Matena,
  Zhou, Li, and Liu}]{t5}
Colin Raffel, Noam Shazeer, Adam Roberts, Katherine Lee, Sharan Narang, Michael
  Matena, Yanqi Zhou, Wei Li, and Peter~J. Liu. 2020.
\newblock Exploring the limits of transfer learning with a unified text-to-text
  transformer.
\newblock \emph{JMLR}, 21:140:1--140:67.

\bibitem[{Schick and Sch{\"{u}}tze(2021)}]{gpt3}
Timo Schick and Hinrich Sch{\"{u}}tze. 2021.
\newblock It's not just size that matters: Small language models are also
  few-shot learners.
\newblock In \emph{NAACL 2021}, pages 2339--2352.

\bibitem[{Schwenk et~al.(2021)Schwenk, Wenzek, Edunov, Grave, Joulin, and
  Fan}]{ccmatrix}
Holger Schwenk, Guillaume Wenzek, Sergey Edunov, Edouard Grave, Armand Joulin,
  and Angela Fan. 2021.
\newblock Ccmatrix: Mining billions of high-quality parallel sentences on the
  web.
\newblock In \emph{ACL 2021}, pages 6490--6500.

\bibitem[{See et~al.(2017)See, Liu, and Manning}]{ptrnet}
Abigail See, Peter~J. Liu, and Christopher~D. Manning. 2017.
\newblock Get to the point: Summarization with pointer-generator networks.
\newblock In \emph{ACL 2017}, pages 1073--1083.

\bibitem[{Song et~al.(2019)Song, Tan, Qin, Lu, and Liu}]{mass}
Kaitao Song, Xu~Tan, Tao Qin, Jianfeng Lu, and Tie{-}Yan Liu. 2019.
\newblock {MASS:} masked sequence to sequence pre-training for language
  generation.
\newblock In \emph{ICML 2019}, volume~97, pages 5926--5936.

\bibitem[{Sun et~al.(2019)Sun, Wang, Li, Feng, Chen, Zhang, Tian, Zhu, Tian,
  and Wu}]{ernie:baidu}
Yu~Sun, Shuohuan Wang, Yukun Li, Shikun Feng, Xuyi Chen, Han Zhang, Xin Tian,
  Danxiang Zhu, Hao Tian, and Hua Wu. 2019.
\newblock {ERNIE}: Enhanced representation through knowledge integration.
\newblock \emph{ArXiv}, abs/1904.09223.

\bibitem[{Tiedemann(2012)}]{tatoeba}
J{\"{o}}rg Tiedemann. 2012.
\newblock Parallel data, tools and interfaces in {OPUS}.
\newblock In \emph{LREC 2012}, pages 2214--2218.

\bibitem[{Vaswani et~al.(2017)Vaswani, Shazeer, Parmar, Uszkoreit, Jones,
  Gomez, Kaiser, and Polosukhin}]{transformer}
Ashish Vaswani, Noam Shazeer, Niki Parmar, Jakob Uszkoreit, Llion Jones,
  Aidan~N. Gomez, Lukasz Kaiser, and Illia Polosukhin. 2017.
\newblock Attention is all you need.
\newblock In \emph{NIPS 2017}, pages 5998--6008.

\bibitem[{Wang et~al.(2019)Wang, Singh, Michael, Hill, Levy, and Bowman}]{glue}
Alex Wang, Amanpreet Singh, Julian Michael, Felix Hill, Omer Levy, and
  Samuel~R. Bowman. 2019.
\newblock {GLUE:} {A} multi-task benchmark and analysis platform for natural
  language understanding.
\newblock In \emph{ICLR 2019}. OpenReview.net.

\bibitem[{Wang et~al.(2020)Wang, Zhai, and Hassan}]{zcode}
Yiren Wang, ChengXiang Zhai, and Hany Hassan. 2020.
\newblock Multi-task learning for multilingual neural machine translation.
\newblock In \emph{EMNLP 2020}, pages 1022--1034.

\bibitem[{Xiao et~al.(2020)Xiao, Zhang, Li, Sun, Tian, Wu, and Wang}]{erniegen}
Dongling Xiao, Han Zhang, Yu{-}Kun Li, Yu~Sun, Hao Tian, Hua Wu, and Haifeng
  Wang. 2020.
\newblock {ERNIE-GEN:} {An} enhanced multi-flow pre-training and fine-tuning
  framework for natural language generation.
\newblock \emph{CoRR}, abs/2001.11314.

\bibitem[{Xue et~al.(2021)Xue, Constant, Roberts, Kale, Al{-}Rfou, Siddhant,
  Barua, and Raffel}]{mt5}
Linting Xue, Noah Constant, Adam Roberts, Mihir Kale, Rami Al{-}Rfou, Aditya
  Siddhant, Aditya Barua, and Colin Raffel. 2021.
\newblock mt5: {A} massively multilingual pre-trained text-to-text transformer.
\newblock In \emph{NAACL 2021}, pages 483--498.

\bibitem[{Yang et~al.(2022{\natexlab{a}})Yang, Huang, Ma, Yin, Dong, Zhang,
  Guo, Li, and Wei}]{crop}
Jian Yang, Shaohan Huang, Shuming Ma, Yuwei Yin, Li~Dong, Dongdong Zhang,
  Hongcheng Guo, Zhoujun Li, and Furu Wei. 2022{\natexlab{a}}.
\newblock {CROP:} zero-shot cross-lingual named entity recognition with
  multilingual labeled sequence translation.
\newblock In \emph{Findings of EMNLP 2022}, pages 486--496.

\bibitem[{Yang et~al.(2021)Yang, Ma, Huang, Zhang, Dong, Huang, Muzio, Singhal,
  Hassan, Song, and Wei}]{microsoft_wmt2021}
Jian Yang, Shuming Ma, Haoyang Huang, Dongdong Zhang, Li~Dong, Shaohan Huang,
  Alexandre Muzio, Saksham Singhal, Hany Hassan, Xia Song, and Furu Wei. 2021.
\newblock Multilingual machine translation systems from microsoft for {WMT21}
  shared task.
\newblock In \emph{WMT 2021}, pages 446--455. Association for Computational
  Linguistics.

\bibitem[{Yang et~al.(2020)Yang, Ma, Zhang, Wu, Li, and Zhou}]{alm}
Jian Yang, Shuming Ma, Dongdong Zhang, Shuangzhi Wu, Zhoujun Li, and Ming Zhou.
  2020.
\newblock Alternating language modeling for cross-lingual pre-training.
\newblock In \emph{AAAI 2020}, pages 9386--9393.

\bibitem[{Yang et~al.(2022{\natexlab{b}})Yang, Yin, Ma, Zhang, Li, and
  Wei}]{hlt_mt}
Jian Yang, Yuwei Yin, Shuming Ma, Dongdong Zhang, Zhoujun Li, and Furu Wei.
  2022{\natexlab{b}}.
\newblock High-resource language-specific training for multilingual neural
  machine translation.
\newblock In \emph{IJCAI 2022}, pages 4461--4467.

\bibitem[{Yang et~al.(2022{\natexlab{c}})Yang, Yin, Ma, Zhang, Wu, Guo, Li, and
  Wei}]{um4}
Jian Yang, Yuwei Yin, Shuming Ma, Dongdong Zhang, Shuangzhi Wu, Hongcheng Guo,
  Zhoujun Li, and Furu Wei. 2022{\natexlab{c}}.
\newblock {UM4:} unified multilingual multiple teacher-student model for
  zero-resource neural machine translation.
\newblock In \emph{IJCAI 2022}, pages 4454--4460.

\bibitem[{Yang et~al.(2019)Yang, Dai, Yang, Carbonell, Salakhutdinov, and
  Le}]{xlnet}
Zhilin Yang, Zihang Dai, Yiming Yang, Jaime~G. Carbonell, Ruslan Salakhutdinov,
  and Quoc~V. Le. 2019.
\newblock {XLNet}: Generalized autoregressive pretraining for language
  understanding.
\newblock In \emph{NeurIPS 2019}, pages 5754--5764.

\bibitem[{Zhang et~al.(2020)Zhang, Williams, Titov, and Sennrich}]{opus_100}
Biao Zhang, Philip Williams, Ivan Titov, and Rico Sennrich. 2020.
\newblock Improving massively multilingual neural machine translation and
  zero-shot translation.
\newblock In \emph{ACL 2020}, pages 1628--1639.

\end{thebibliography}
\bibliographystyle{acl_natbib}

\clearpage
\appendix

\appendix
\section{Statistics of Datasets}

\paragraph{WMT-14 En-De} WMT-14 En-De consists of 4.5M sentence pairs. The validation set is devtest2014, and the test set is newstest2014.\footnote{\url{https://statmt.org/wmt14/translation-task.html}}

\paragraph{WMT-16 En-Fr} WMT-14 En-Fr is a large-scale dataset containing nearly 41M sentence pairs, where newstest2014 is employed for evaluation.

\paragraph{WMT-16 En-Ro} WMT-16 En-Ro is comprised of original parallel sentences and back-translation data. We use newsdev2016 for validation and newstest2016 for test. Following the previous work \cite{mbart}, we use the same back-translation data for a fair comparison.\footnote{\url{https://www.statmt.org/wmt16/translation-task.html}}

\paragraph{IWSLT-2017} We download English (En), German (De), Italian (It), Dutch (Nl), and Romanian (Ro) corpora from the IWSLT-2017 benchmark. The dev2010 is used for validation and tst2017 for test.\footnote{\url{https://sites.google.com/site/iwsltevaluation2017/TED-tasks}}

\paragraph{WMT-10} Table~\ref{tab:wmt10} lists the detailed statistics of 10 language pairs from WMT-10, which is a collection of parallel data in different languages from the WMT datasets.
The parallel data is paired with English and other 10 languages, including French (Fr), Czech (Cs), German (De), Finnish (Fi), Latvian (Lv), Estonian (Et), Romanian (Ro), Hindi (Hi), Turkish (Tr) and Gujarati (Gu). The corpora of the WMT benchmark, exclude WikiTiles, from the latest available year of each language are chosen. After removing the duplicated samples, we limit the size of each parallel language pair data up to 10 million by randomly sampling from the whole corpus. We adopt the same valid and test sets from the WMT benchmark as the previous work \cite{zcode}.

\paragraph{WikiLingua} To test the capability of our multilingual pre-trained model, a large-scale multilingual dataset named \textbf{WikiLingua} \cite{wikilingua} of 18 languages from WikiHow is used to evaluate multilingual abstractive summarization systems.\footnote{\url{https://github.com/esdurmus/Wikilingua}}

\begin{table}[htb]
\centering
\resizebox{1.0\columnwidth}{!}{
\begin{tabular}{ccccccc}
\toprule
Code & Language & \#Bitext & Training & Valid & Test \\
\midrule
Fr & French & 10M     & WMT15 & Newstest13 & Newstest15 \\
Cs & Czech & 10M      & WMT19 & Newstest16 & Newstest18 \\
De & German & 4.6M    & WMT19 & Newstest16 & Newstest18 \\
Fi & Finnish & 4.8M   & WMT19 & Newstest16 & Newstest18 \\
Lv & Latvian & 1.4M   & WMT17 & Newsdev17 & Newstest17 \\
Et & Estonian & 0.7M  & WMT18 & Newsdev18 & Newstest18 \\
Ro & Romanian & 0.5M  & WMT16 & Newsdev16 & Newstest16 \\
Hi & Hindi & 0.26M    & WMT14 & Newsdev14 & Newstest14 \\
Tr & Turkish & 0.18M  & WMT18 & Newstest16 & Newstest18 \\
Gu & Gujarati & 0.08M & WMT19 & Newsdev19 & Newstest19 \\
\bottomrule
\end{tabular}}
\caption{Statistics and sources of the training, valid, and test sets from WMT between English and other languages.}
\vspace{-10pt}
\label{tab:wmt10}
\end{table}

\section{Pre-training and Fine-tuning Details}
\paragraph{Pre-training Hyper-parameters} Table \ref{tab:pretraining_hyperparams} summarizes the hyper-parameters for pre-training \ourmethod{} and \mourmethod{}
The task-specific hyper-parameters for the downstream language generation and understanding tasks are in Table \ref{tab:task_specific_hyperparams}. 

\paragraph{Abstractive Summarization} 
During fine-tuning, we use the Adam \cite{adam} optimizer with an initial learning rate of 1e-4 and the batch size is set as 2048 tokens on 8 V100 GPUs. The models
are trained with the label smoothing cross-entropy with a smoothing ratio of 0.1. The last 5 checkpoints are averaged for evaluation. 

\paragraph{Neural Machine Translation} We adopt Adam with a learning rate of 1e-4 and set the batch size as 2048 tokens on 8 V100 GPUs for all bilingual translation tasks and the IWSLT-2017 benchmark. For the large-scale multilingual dataset WMT-10, our pre-trained model is fine-tuned on 32 V100 GPUs with a learning rate of 3e-4. For a fair comparison, we adopt the same architecture and model size as our pre-trained model.
\paragraph{Data-to-text Generation} 
We use Adam with a learning rate of \{8e-5,1e-4\} and set the batch size as 16 sentences on the WebNLG dataset.

\paragraph{Multi-lingual Fine-tuning} Following the previous work~\cite{zcode,deltalm}, we adopt a dynamic temperate-based sampling strategy to mitigate the unbalance of the multilingual corpora, where we gradually sample more pairs in low-resource languages with the number of epochs increasing. The temperature of the $i$-th epoch is calculated by:
\begin{equation}
\begin{MiddleEquation}
    \tau_{i}=\min(\tau_{1}, \tau_{0}+\frac{i}{N}(\tau-\tau_{0}))
    \label{epoch_temperature}
\end{MiddleEquation}
\end{equation}
where $\tau_{0}$ is the initial temperature, $\tau_{1}$ is the peak temperature, and $N$ is the number of warm-up epochs. We set $\tau_{0}=1.0$, $\tau_{1}=5.0$, and $N=5$ for all multilingual experiments for a fair comparison. 

Given the temperature $\tau_{i}$ $i$-th epoch, we can calculate the real sampling ratio of the language $L_{k}$, where $L_{k} \in L_{all}=\{L_{1},\dots,L_{K}\}$:
\begin{equation}
\begin{MiddleEquation}
    q_{L_{k}}(i)=\frac{p_{L_{k}}^{\frac{1}{\tau_{i}}}}{\sum_{j=1}^{K} p_{L_{j}}^{\frac{1}{\tau_{i}}}}
\end{MiddleEquation}
\end{equation}
where $q_{L_{k}}(i)$ is the sampling ratio of the language $L_k$ in the $i$-th epoch. $p_{L_k}$ is the real data ratio of the language $L_{k}$ in all languages. $\tau_{i}$ is the temperature of the $i$-th epoch, as described in Equation \ref{epoch_temperature}.

\begin{table}[t]
\begin{center}
\resizebox{1.0\columnwidth}{!}{
\begin{tabular}{lccc}
\toprule
\bf Hyper-parameter  & \bf \ourmethod{} & \bf \mourmethod{} \\
\midrule 
Number of Encoder Layers             & 12 & 12 \\
Number of Generator Layers           & 12 & 12 \\
Number of Discriminator Layers       & 4 & 4 \\
Hidden size                          & 768 & 768 \\
FFN hidden size                      & 3072 & 3072 \\
Attention heads        & 12 & 12 \\
Attention head size    & 64 & 64 \\
Dropout                & 0.1 & 0.1 \\
Attention Dropout      & 0.1 & 0.1 \\
Warmup Steps           & 10k & 10k \\
Peak Learning Rate     & 4e-4 & 5e-4 \\
Batch Size             & 8K & 8K \\
Weight Decay           & 0.01 & 0.01 \\
Max Steps              & 500k & 500k\\
Learning Rate Decay    & Linear & Linear \\
Adam $\beta_1$         & 0.9 & 0.9 \\
Adam $\beta_2$         & 0.98 & 0.98 \\
Gradient Clipping      & 0.0 & 0.0 \\
\bottomrule
\end{tabular}}
\end{center}
\caption{Hyper-parameters for pre-training \ourmethod{} and \mourmethod{}.}
\label{tab:pretraining_hyperparams}
\end{table}

\begin{table*}[htb]
\resizebox{1.0\textwidth}{!}{
\centering
\begin{tabular}{lccccccccc}
\toprule
\bf Task &\bf Learning Rate & \bf Warmup Steps & \bf Batch Size  & \bf Weight Decay & \bf Max Epoch & \bf Gradient Clipping & \bf \makecell[c]{Max Source Positions} & \bf \makecell[c]{Max Target Positions} \\
\midrule
\multicolumn{6}{l}{\textit{Text Summarization}} \\
\midrule
CNN / DailyMail & 1e-4 & 1000  & 2048 (Tokens) & 0.0 & 16 & 0.0 & 608 & 160  \\
XSum            & 1e-4 & 1000  & 2048 (Tokens) & 0.0 & 16 & 0.0 & 720 & 48 \\
WikiLingua     & 1e-4 & 1000  & 2048 (Tokens) & 0.0 & 16 & 0.0 & 512 & 160 \\
\midrule
\multicolumn{6}{l}{\textit{Machine Translation}} \\
\midrule
WMT14 En-De     & 1e-4 & 4000  & 2048 (Tokens) & 0.0 & 50  & 0.0 & 512 & 512\\
WMT14 En-Fr     & 1e-4 & 4000  & 2048 (Tokens) & 0.0 & 50  & 0.0 & 512 & 512\\
WMT14 En-Ro     & 1e-4 & 4000  & 2048 (Tokens) & 0.0 & 16  & 0.0 & 512 & 512\\
IWSLT17         & 1e-4  & 4000  & 2048 (Tokens) & 0.05 & 16 & 0.0 & 512 & 512 \\
WMT10           & 3e-4  & 4000  & 2048 (Tokens) & 0.0 & 8  & 0.0 & 512 & 512 \\
\midrule
\multicolumn{6}{l}{\textit{Data-to-Text}} \\
\midrule
WebNLG          & \{2.5e-5, 5e-5\}  & 1000  & 2048 (Tokens) & 0.05 & 16 & 0.0 & 512 & 512 \\
\midrule
\multicolumn{6}{l}{\textit{Natural Language Understanding}} \\
\midrule
XNLI          & \{2.5e-5, 5e-5\}  & 4000  & 16 (Sentences) & 0.05 & 30 & 1.0 & 512 & 512 \\
GLUE          & \{1e-5, 2.5e-5, 5e-5\}  & 4000  & \{8,16\} (Sentences) & 0.05 & 30 & 1.0 & 512 & 512 \\
\bottomrule
\end{tabular}
}
\caption{
Task-specific hyper-parameters for downstream language generation and understanding benchmarks.
}
\label{tab:task_specific_hyperparams}
\end{table*}

\section{Results on Downstream Task}
\paragraph{GLUE} For each classification task of the GLUE \cite{glue}, we conduct 5 experiments with different seeds $\{1,2,3,4,5\}$ and report the average accuracy of 5 experiments.

\paragraph{XNLI} We also conduct 5 experiments with different seeds $\{1,2,3,4,5\}$ and report the average accuracy of 5 experiments.

\begin{figure*}[t]
\begin{center}
	\includegraphics[width=1.0\textwidth]{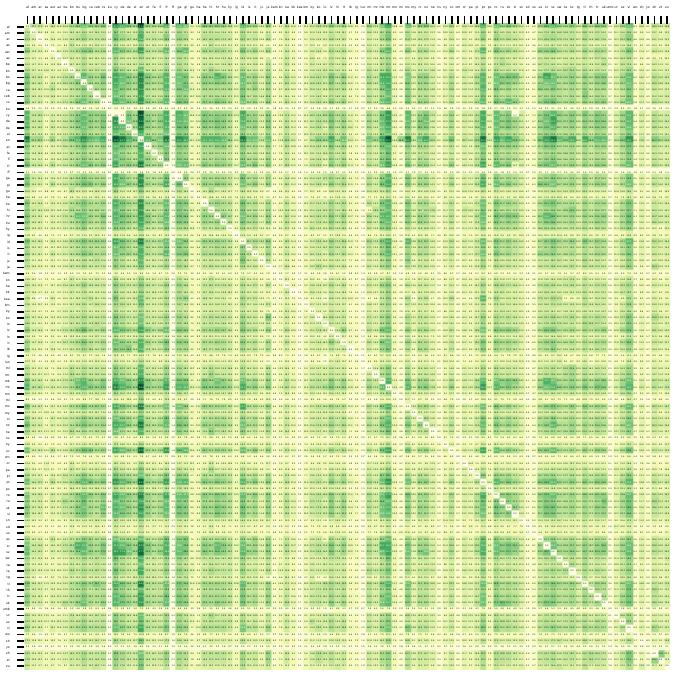}
	\caption{Evaluation results of our multilingual model on all translation directions on the FLORES-101 devtest set, where our model consists of 12 encoder and 12 decoder layers with a hidden size of 768. We fine-tune the multilingual encoder-decoder pre-trained model \mourmethod{} on the large-scale dataset.
	The language $x$ in the $i$-th row and language $y$ in the $j$-th column denotes the translation direction from the language $x$ to language $y$. For example, the cell of the $1$-st row (af) and the $3$-nd column (ar) represents the result of the translation direction af$\to$ar. The table shows the results of all translation directions of 102 languages.}
	\label{flores_all}
\end{center}
\end{figure*}

\paragraph{FLORES} Since the corpora of $X \to Y$ are commonly scarce, the performance of low-resource translation direction Avg$_{X \to Y}$ mainly depends on the zero-shot cross-lingual transferability of the pre-trained model. Our model with the 12 encoder layers and 12 decoder layers significantly outperforms the previous state-of-the-art model M2M-124 with large model size.
In Figure \ref{flores_all}, we report the multilingual model initialized by our pre-trained model in all translation directions, where the languages are ordered alphabetically by the language code.
Following the previous work \cite{microsoft_wmt2021}, we use the same training data, including CCAligned \cite{ccalign}, CCMatrix \cite{ccmatrix}, OPUS-100 \cite{opus_100}, JW300 \cite{jw300}, Tatoeba \cite{tatoeba}, WMT2021 news track\footnote{\url{http://statmt.org/wmt21/translation-task.html}}, multilingual track data\footnote{\url{http://data.statmt.org/wmt21/multilingual-task/}}.
\begin{table}[t]
\centering
\resizebox{0.9\columnwidth}{!}{
\begin{tabular}{lccccccc}
\toprule
 \bf Seed     & \bf MNLI & \bf SST-2 & \bf MRPC & \bf RTE & \bf QNLI  & \bf QQP & \bf Avg$_{\bm{6}}$ \\ \midrule
\multicolumn{8}{l}{\textit{Fine-tuning on Discriminator ($\mathcal{D}$)}}    \\
 \midrule
1        & 88.9   & 94.5  & 89.7  & 83.8   & 93.8  & 91.6     & 90.4     \\
2        & 89.1   & 94.7  & 90.0  & 84.8   & 93.9  & 91.7     & 90.7     \\
3        & 88.9   & 94.5  & 91.7  & 83.0   & 93.7  & 91.9     & 90.6     \\
4        & 89.0   & 94.7  & 90.9  & 84.1   & 93.8  & 91.8     & 90.7     \\
5        & 89.2   & 95.2  & 90.7  & 80.1   & 94.2  & 91.7     & 90.2     \\
Avg      & 89.0   & 94.7  & 90.6  & 83.2   & 93.9  & 91.7     & 90.5     \\
 \midrule
\multicolumn{8}{l}{\textit{Fine-tuning on Generator ($\mathcal{G}$)}} \\
 \midrule
1       & 89.2   & 95.1   & 90.4  & 85.6   & 94.1  & 91.9     & 91.0    \\
2       & 89.1   & 95.2   & 90.9  & 85.6   & 94.3  & 92.1     & 91.2    \\
3       & 89.2   & 95.0   & 90.4  & 84.5   & 94.1  & 91.9     & 90.9    \\
4       & 89.4   & 95.1   & 90.9  & 84.8   & 94.1  & 92.1     & 91.1    \\
5       & 89.6   & 94.8   & 89.7  & 84.5   & 94.2  & 91.8     & 90.8    \\
Avg     & 89.3   & 95.0   & 90.5  & 85.0   & 94.2  & 92.0     & 91.0    \\
\bottomrule
\end{tabular}}
\caption{The accuracy scores of the base-setting models on the valid set of GLUE classification tasks.}
\label{tab:glue_seed}
\vspace{-10pt}
\end{table}

\begin{table*}[htb]
\centering
\resizebox{1.0\textwidth}{!}{
\begin{tabular}{l|ccccccccccccccc|c}
\toprule
\bf Model & \bf En & \bf Ar & \bf Bg & \bf De & \bf El & \bf Es & \bf Fr & \bf Hi & \bf Ru & \bf Sw & \bf Th & \bf Tr & \bf Ur & \bf Vi & \bf Zh & \bf Avg$_{\bm{15}}$ \\
\midrule
\multicolumn{17}{l}{\emph{Cross-lingual zero-shot transfer (models fine-tune on English data only)}} \\
\midrule

mBERT     & 80.8 & 64.3& 68.0& 70.0& 65.3& 73.5& 73.4& 58.9& 67.8& 49.7& 54.1& 60.9& 57.2& 69.3& 67.8& 65.4 \\
XLM       & 85.0 & 73.1 & 77.4 & 77.8 & 76.6  & 78.9 & 78.7 & 69.6 & 75.3 & 68.4 & 73.2 & 72.5 &67.3 &76.1 &76.5 & 75.1\\
mT5-Small & 79.6 & 65.2 & 71.3 & 69.2 & 68.6  & 72.7 & 70.7 & 62.5 & 70.1 & 59.7 & 66.3 & 64.4 &59.9 &66.3 & 65.8 & 67.5 \\
mT5-Base  & 84.7 & 73.3 & 78.6 & 77.4 & 77.1  & 80.3 & 79.1 & 70.8 & 77.1 & 69.4 & 73.2 & 72.8 & 68.3 & 74.2 & 74.1 & 75.4 \\
\mourmethod{} (D) &     85.9 &     72.6 & \bf 78.6 &     78.6 & \bf 76.6 & \bf 80.7 &     79.8 &     70.4 &     76.0 & \bf 64.4 & \bf 74.3 &     74.4 &     66.5 & \bf 77.2 & \bf 75.9 & \bf 75.5\\
\mourmethod{} (G) & \bf 86.3 & \bf 73.2 &     77.9 & \bf 79.0 &     76.5 &     80.3 & \bf 80.4 & \bf 70.8 & \bf 76.7 &     62.9 &     74.2 & \bf 74.5 & \bf 66.6 &     76.5 &     75.7 &     75.4\\
\midrule
\multicolumn{17}{l}{\emph{Translate-train (models fine-tune on English training data plus translations in all target languages)}} \\
\midrule
XLM & 85.0  &76.5 &79.3 & 80.3 & 78.1 & 80.3&80.2& 72.3&78.1& 70.9&75.5&74.7 &63.2&76.6&78.6& 76.6  \\
\mourmethod{} (D)&      85.9& \bf 76.9  & \bf 79.9 &     80.7  & \bf 79.5 & \bf 81.6 &     80.9 &     74.2 & \bf 78.7 & \bf 71.8 &     76.9 &     76.9 & \bf 65.8 &     79.1 & \bf 80.0 & 77.9 \\
\mourmethod{} (G)&  \bf 86.3&     76.7  &     79.7 & \bf 80.8  &     79.7 &     81.6 & \bf 82.0 & \bf 74.6 &     78.6 &     70.8 & \bf 77.4 & \bf 77.1 &     65.3 & \bf 79.2 &    79.3  & \bf 77.9 \\
\midrule
\multicolumn{17}{l}{\emph{Translate-train (models fine-tune on English training data plus translations in all target languages)}} \\
\midrule
XLM       & 85.0  & 77.6 & 80.9 & 80.3 & 79.1 & 81.3 & 80.8 & 72.9 & 78.3 & 72.8 & 76.0 & 75.6 & 68.5 & 78.5 & 79.5 & 77.8  \\
mT5-Small & 69.5  & 63.7 & 67.5 & 65.7 & 66.4 & 67.5 & 67.3 & 61.9 & 66.4 & 59.6 & 63.9 & 63.5 & 60.4 & 63.3 & 64.5 & 64.7 \\
mT5-Base & 82.0 & 74.4 & 78.5 & 77.7 & 78.1 & 79.1 & 77.9 & 72.2 & 76.5 & 71.5 & 75.0 & 74.8 & 70.4 & 74.5 & 76.0   & 75.9 \\
\mourmethod{} (D)&  \bf 87.3 & \bf 78.3 &     82.7 & \bf 83.1 &     82.2 &     83.8 &     83.3 & \bf 77.3 &     81.3 &     73.1 & \bf 80.3 & \bf 79.9 &     71.2 &     81.3 & \bf 81.8 &     80.5 \\
\mourmethod{} (G)&      87.2 &     78.3 & \bf 83.3 &     82.7 & \bf 82.3 & \bf 84.0 & \bf 83.6 &     77.1 & \bf 81.4 & \bf 74.5 &     79.8 &     79.6 & \bf 71.3 & \bf 81.6 &     81.6 & \bf 80.6 \\
\bottomrule
\end{tabular}}
\caption{XNLI accuracy scores for each language.}
\end{table*}

\section{Weight Sharing}
\begin{table}[t]
\centering
\resizebox{1.0\columnwidth}{!}{
\begin{tabular}{c|c|c|c|c}
\toprule
\bf ID      & \bf \#Params   & \bf Strategy   &  \makecell[c]{\textbf{Xsum} \\ RG-1/RG-2/RG-L} & \makecell[c]{\textbf{WMT16 En-Ro} \\ En$\to$Ro/Ro$\to$En}    \\
\midrule
{\large{\ding{172}}} & 390M &$\theta_{\mathcal{G}} =\theta_{\mathcal{D}}$& 43.26/19.82/35.02     &  37.4/37.2\\
{\large{\ding{173}}} & 430M &$\theta_{\mathcal{G}}\neq \theta_{\mathcal{D}}$& \bf 45.36/21.98/36.84 &\bf 38.3/38.0 \\ 

\bottomrule
\end{tabular}
}
\caption{Evaluation results with different weight sharing strategies on the test set of the Xsum summarization task and WMT16 En-Ro translation task. Both generator decoder $\theta_{\mathcal{G}}$ and discriminator decoder $\theta_{\mathcal{D}}$ have 12 layers in Experiment {\large{\ding{173}}} by sharing decoder parameters.
}
\label{weight_sharing}
\vspace{-10pt}
\end{table}

Our pre-trained model includes the discriminator $(\mathcal{D}: \{\theta_{\mathcal{E}},\theta_{\mathcal{D}}\})$ and generator $(\mathcal{G}: \{\theta_{\mathcal{E}},\theta_{\mathcal{G}}\})$. We can use a 12-layer generator decoder $\theta_{\mathcal{G}}$ and a 4-layer tiny discriminator decoder $\theta_{\mathcal{D}}$ for \gtask{}. We propose a weight sharing strategy to improve the model efficiency of the pre-training by sharing weights among the generator and decoder ($\theta_{\mathcal{D}}=\theta_{\mathcal{G}}$) by setting the discriminator generator and generator decoder as the same size (both 12 layers). Table \ref{weight_sharing} lists the results of different weight sharing strategies. It turns out the sharing decoder setting performs worse than not sharing. It is reasonable since the generator decoder is used for sequence generation whereas the discriminator decoder is a classifier.

\end{document}